\newif\ifconfver
\newif\ifonecoltab
\newif\ifplainver  %declare a plain version
\plainvertrue%%

\ifplainver
    \confverfalse   %automatically disable conf. version argument if it's plain
\fi

\ifconfver
     \documentclass[10pt,twocolumn,twoside]{IEEEtran}
\else
    \ifplainver
        \documentclass[11pt]{article}
        \usepackage{fullpage}
    \else
        \documentclass[12pt,draftcls,onecolumn]{IEEEtran}
    \fi
\fi

\usepackage{calc,amsfonts,amssymb,amsmath,bm,url,color,theorem,graphicx,cite}
\usepackage{psfrag,subfigure,float}
\usepackage[algoruled,linesnumbered]{algorithm2e}
\usepackage{multirow}
\definecolor{orange}{RGB}{255,107,0}

\newtheorem{Lemma}{Lemma}

\newtheorem{Claim}{Claim}
\theorembodyfont{\rmfamily}

\newtheorem{Remark}{Remark}

\begin{document}

\bibliographystyle{IEEEtran}

%--- I do things quite strangely here to accommodate three style modes.
%--- input title and abstract here; it applies to all modes
%--- it's too complex to do authors or they are input for each mode
\newcommand{\papertitle}{
{Joint Tensor Factorization and Outlying Slab Suppression with Applications}
}

\newcommand{\paperabstract}{
We consider factoring low-rank tensors in the presence of outlying slabs.
This problem is important in practice, because data collected in many real-world applications, such as speech, fluorescence, and some social network data, fit this paradigm.
Prior work tackles this problem by iteratively selecting a fixed number of slabs and fitting, a procedure which may not converge.
We formulate this problem from a group-sparsity promoting point of view, and propose an alternating optimization framework to handle the corresponding $\ell_p$ ($0<p\leq 1$)
minimization-based low-rank tensor factorization problem.
The proposed algorithm features a similar per-iteration complexity as the plain trilinear alternating least squares (TALS) algorithm.
Convergence of the proposed algorithm is also easy to analyze under the framework of alternating optimization and its variants.
In addition, regularization and constraints can be easily incorporated to make use of \emph{a priori} information on the latent loading factors.
Simulations and real data experiments on blind speech separation, fluorescence data analysis, and social network mining are used to showcase the effectiveness of the proposed algorithm.
}

%--------

\ifplainver

    %\date{May 30, 2014}

    \title{\papertitle}

    \author{
    $^\dag$Xiao Fu, $^\dag$Kejun Huang, $^*$Wing-Kin Ma, $^\dag$Nicholas D. Sidiropoulos, and $^\ddag$Rasmus Bro
    \\ ~ \\
		$^\dag$Department of Electrical and Computer Engineering, University of Minnesota,\\
		Minneapolis, 55455, MN, United States\\
		Email: (xfu,huang663,nikos)@umn.edu
		\\~\\
    $^*$Department of Electronic Engineering, The Chinese University of Hong  Kong, \\
    Hong Kong \\
    Email: wkma@ieee.org
		\\~\\
		$^\ddag$Department of Food Science, Faculty of Science, University of Copenhagen,\\
		Rolighedsvej 30, DK-1958 Frederiksberg C\\
		Email: rb@life.ku.dk
    }

    \maketitle

    \begin{abstract}
    \paperabstract
    \end{abstract}

\else
    \title{\papertitle}

    \ifconfver \else {\linespread{1.1} \rm \fi
    \author{Xiao Fu, Kejun Huang, Wing-Kin Ma, Nicholas D. Sidiropoulos,~\IEEEmembership{Fellow,~IEEE}, and Rasmus Bro

   	\thanks{Submitted to {\em IEEE Trans. on Signal Processing}, April 3, 2015; revised \today. X. Fu, K. Huang and N.D. Sidiropoulos were supported in part by NSF IIS-1247632.}
    \thanks{X. Fu, K. Huang, and N.D. Sidiropoulos are with the Department of Electrical and Computer Engineering, University of Minnesota, Minneapolis, email: (xfu,huang663,nikos)@umn.edu. W.K. Ma is with the Department of Electronic Engineering, the Chinese University of Hong Kong, Hong Kong, email: wkma@ieee.org.
		R. Bro is with the Department of Food Science, Faculty of Science, University of Copenhagen, Rolighedsvej 30, DK-1958 Frederiksberg C, email: rb@life.ku.dk}
		
		}

    \maketitle

    \ifconfver \else
        \begin{center} \vspace*{-2\baselineskip}
        %11th Revision, \today \\[2\baselineskip]
        \end{center}
    \fi

    \begin{abstract}
    \paperabstract
    \end{abstract}

    \begin{keywords}\vspace{-0.0cm}
        PARAFAC, canonical polyadic decomposition, tensor decomposition, outliers, robustness, iteratively reweighted, group sparsity
    \end{keywords}

    \ifconfver \else \IEEEpeerreviewmaketitle \fi

 \fi

\ifconfver \else
    \ifplainver \else
        \newpage
\fi \fi
%---------------------------------------------------------------------------

\clearpage
\vspace*{-0.1in}
\section{Introduction}
Factoring a tensor (i.e., a data set indexed by three or more indices) into rank-one components is a decomposition problem which is frequently referred to as
\emph{parallel factor analysis} (PARAFAC) or \emph{canonical decomposition} (CANDECOMP), or \emph{canonical polyadic decomposition} (CPD).
Unlike two-way factor analysis (i.e., matrix factorizations), three- or higher-way low-rank tensor factorization reveals essentially unique factors under quite mild conditions,
which is desirable when dealing with latent parameter estimation problems.
Since the late 1990s, PARAFAC has been successfully applied to wireless communications for blindly estimating the spatial channels or the users' code-division signatures \cite{sidiropoulos2000blind,RonVorGerSid:2005}; array processing for finding the directions-of-arrival of the emitters \cite{Sidiropoulos2002,Sidiropoulos2001}; chemometrics for resolving the spectra of chemical analytes \cite{smilde2005multi}; blind speech and audio separation for estimating the mixing system \cite{Nion2010,Reilly2005};
and, more recently, power spectra separation for cognitive radio \cite{fu2014tensor}, and big data mining for social group clustering \cite{papalexakis2013large}.

A high-order tensor can also be considered as a set of lower-order tensors.
For example, a data cube (i.e., a three-way tensor) can be considered as a set of matrices (two-way tensors), obtained by fixing one index to a particular value.
Each such piece of the original data, whose order has been reduced by one, will be called a \emph{slab}.
%Such a set of tenors with one less dimension of the original tensor are called \emph{slabs} or slices of the multi-way data.
Slabs are usually physically meaningful in various applications.
For example, in blind speech and audio separation, the received signals' short-term covariance matrix, assumed constant within a short coherence interval and sometimes referred to as {\em local covariance} \cite{lee2013khatri,FuMaHuaSid2015}, can be considered as a slab of a three-way tensor;
in fluorescence data spectroscopy, a measurement matrix that consists of emissions and excitations of the stimulated analytes is a slab \cite{smilde2005multi};
and in array processing, the received raw signals at a subarray can be considered as a slab \cite{Sidiropoulos2002}.
Due to this physical correspondence, however, strong data contamination or corruption frequently happens at the slab level (rather than element-wise).
A typical example is blind speech separation - it has been observed that locally correlated speech sources may create local covariances (slabs) that do not obey the low-rank tensor model \cite{FuMaHuaSid2015}.
%Also, in fluorescence spectroscopy, the Rayleigh and Raman scattering frequently generate badly contaminated slabs \cite{engelen2011detecting,hubert2012robust,bro2011eemizer}.
%Directly applying existing PARAFAC algorithms to these problems without considering the corrupt slabs may yield poor results.
Also, in chemometrics, e.g. in fluorescence spectroscopy, it is common that certain samples representing erratic measurements or samples of unusual constitution end up influencing the fitted model badly \cite{engelen2011detecting,hubert2012robust,bro2011eemizer}.

Factoring a low-rank tensor in the presence of outlying slabs has been considered before.
In the literature, the most closely related work may be \cite{engelen2011detecting}.
There, an algorithm that iteratively selects a fixed number of slabs to fit with a low-rank tensor model was proposed.
A main drawback with this algorithm is that it may not converge. Also, it is not easy to determine how many slabs should be selected to fit in advance.
Similar insights are also seen in the analytic chemistry context; see \cite{CEM:CEM1208,hubert2012robust,bro2011eemizer}.
In \cite{L1-PARAFAC}, the authors considered a different yet related scenario. There, a PARAFAC approach was proposed by changing the least squares-based optimization criterion to the $\ell_1$-norm based fitting criterion, to make the low-rank decomposition robust against outlying elements.
The resulting algorithms are \emph{alternating linear programming} or \emph{alternating weighted median filtering} (WMF).
The algorithms in \cite{L1-PARAFAC} do not need to pre-define the number of slabs to select for fitting, but they
can be inefficient even when the problem size is medium.
In addition, the $\ell_1$ criterion is optimal in the maximum likelihood sense, when the noise follows the i.i.d. Laplacian distribution;
but it is not specialized for (strong) slab-level outliers, as will be shown in the simulations.

\bigskip

\noindent
\textbf{Contributions:}~In this work, we consider modeling and tackling the low-rank tensor decomposition problem with outlying slabs from a different perspective.
%Our major contribution lies in the algorithm.
Specifically, we formulate the problem from a group-sparsity promoting viewpoint, and come up with an $\ell_p$ ($0<p\leq 1$) fitting criterion.
We propose to tackle this hard optimization problem using an alternating optimization strategy:
by judiciously recasting the original problem into a more convenient form,
we show that it can be tackled using a simple algorithm whose block updates admit closed-form solutions.
This algorithm tends to iteratively select some clean slabs to fit with a PARAFAC model and downweight the outlying slabs at the same time.
Reminiscent of classical robust fitting, the proposed algorithm does not assume knowledge of the number of clean slabs.
Plus, drawing from existing theoretical results on alternating optimization \cite{bertsekas1999nonlinear} and its variants such as \emph{maximum block improvement} (MBI) \cite{MBI}, convergence of the proposed algorithm can be characterized.
It is also worth noting that the proposed algorithm has almost the same per-iteration complexity as the \emph{trilinear alternating least squares} (TALS) algorithm \cite{sidiropoulos2000blind,RonVorGerSid:2005,Sidiropoulos2002}, which is computationally much cheaper than the algorithms in \cite{L1-PARAFAC}.

Extensions to regularized and constrained cases are also considered in this work, since incorporating \emph{a priori} information on the loading factors is important in applied data analysis. Following the same alternating optimization framework, we propose to handle the subproblems by employing an \emph{alternating direction method of multipliers} \cite{Boyd11} (ADMM)-based algorithm,
which allows us to deal with different types of regularization and constraints of interest, under a unified update strategy.

Besides simulations using synthetic data for verifying the ideas,
we use several simulations and experiments with real data to showcase the effectiveness of the proposed approaches.
First, the basic robust algorithm is applied on blind speech separation simulations, where real speech segments are mixed under realistic room acoustic impulse response scenarios. The separation performance of the proposed algorithm is shown to be superior to the earlier state-of-the-art.
Then, the proposed algorithm is applied to a fluorescence data set; and finally to the ENRON e-mail corpus.
Interesting and nicely interpretable results are obtained in both cases.

\noindent \emph{Notation}:
We largely follow standard signal processing (and some Matlab) notational conventions, for convenience.
Specifically,
$\underline{\bf X}\in\mathbb{R}^{I\times J\times K}$ denotes a three-way tensor, and
$\underline{\bf X}(i,j,k)$ denotes the element that is indexed by $(i,j,k)$;
$\underline{\bf X}(i,:,:)$, $\underline{\bf X}(:,j,:)$
and $\underline{\bf X}(:,:,k)$ denote the $i$th horizontal slab, the $j$th lateral slab, and the $k$th frontal slab, respectively;
${\bf X}(i,:)$ and ${\bf X}(:,j)$ denotes the $i$th row and the $j$th column of the matrix ${\bf X}$;
$^T$ denotes the transpose operator;
$^\dag$ denotes the Moore–Penrose pseudo-inverse operator;
$\|{\bf x}\|_p = (\sum_{i=1}^m |x_i|^p)^{1/p}$ for ${\bf x}\in\mathbb{R}^{m}$ for $0<p<\infty$;
${\rm vec}({\bf X})=[{\bf X}^T(:,1),\ldots,{\bf X}^T(:,J)]^T$ for ${\bf X}\in\mathbb{R}^{I\times J}$;
${k}_{\bf X}$ and ${\rm rank}({\bf X})$ denote the Kruskal rank and the rank of ${\bf X}$, respectively;
$\circ$, $\circledast$, $\otimes$ and $\odot$ denote
the outer product, the Hadmard product, the Kronecker product, and the Khatri-Rao product, respectively;
${\rm Diag}({\bf x})$ denotes a diagonal matrix that holds the $x_1,\ldots,x_m$ as the diagonal elements.

\vspace*{-0.1in}
\section{Preliminaries on PARAFAC}\label{sec:preliminaries}
\begin{figure}[!h]
	\centering
	  \psfrag{N}[B]{\small $I$}
		 \psfrag{F}[B]{\small $J$}
		 \psfrag{L}[B]{\small $K$}
		 \psfrag{G}[B]{\small $\underline{\bf X}$}
		 \psfrag{G1}[B]{\small ${\bf X}^{(2)}_k={\bf A}{\bf D}_k({\bf C}){\bf B}^T$}
		  \psfrag{G2}[B]{\small ${\bf X}^{(3)}_i={\bf  B}{\bf D}_i({\bf A}){\bf C}^T$}
	   \psfrag{G3}[B]{\small ${\bf X}^{(1)}_j={\bf C}{\bf D}_j({\bf B}){\bf A}^T$}
		\includegraphics[width=8cm]{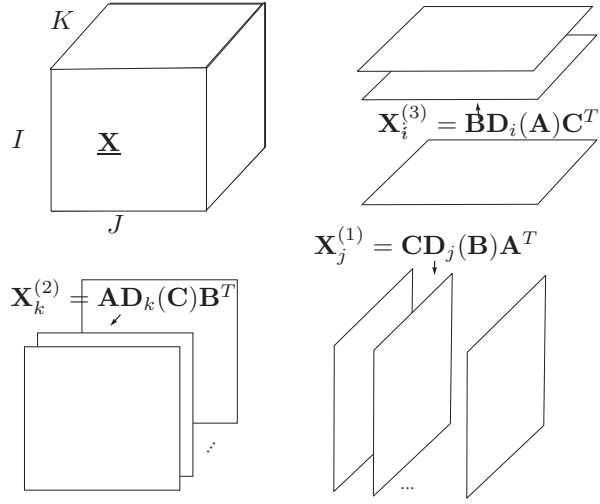}
	\caption{Slabs of a three-way tensor.}
	\label{fig:tensor_1}
\end{figure}

A simple description of the PARAFAC model is as follows. PARAFAC aims to represent a three-way tensor $\underline{\bf X}\in\mathbb{C}^{I\times J \times K}$ using
PARAFAC three latent factor matrices:
\begin{equation}
\underline{\bf X}\approx \sum_{r=1}^R{\bf A}(:,r)\circ{\bf B}(:,r)\circ{\bf C}(:,r),
\label{eq:PARAFAC}
\end{equation}
where ${\bf A}\in\mathbb{C}^{I\times R}$, ${\bf B}\in\mathbb{C}^{J\times R}$, ${\bf C}\in\mathbb{C}^{K\times R}$, and $R$ is called the rank of the PARAFAC model. Any tensor $\underline{\bf X}\in\mathbb{C}^{I\times J \times K}$ can be exactly represented this way if a large-enough $R \leq \min(IJ,JK,IK)$ is used; but we are usually interested in using relatively small $R$ to capture the `principal components' of $\underline{\bf X}$.
Equivalently, each element of the tensor can be represented as $\underline{\bf X}(i,j,k)\approx\sum_{r=1}^R{\bf A}(i,r){\bf B}(j,r){\bf C}(k,r)$.
A three-way tensor is also a set of matrices, or, slabs,
which are obtained by fixing one index.
There are three types of slabs of a three-way tensor, namely, the horizontal slabs ($\{\underline{\bf X}(i,:,:)\}_{i=1}^I$), the lateral slabs ($\{\underline{\bf X}(:,j,:)\}_{j=1}^J$), and the frontal slabs ($\{\underline{\bf X}(:,:,k)\}_{k=1}^K$).
If the PARAFAC model in \eqref{eq:PARAFAC} holds exactly, each type of slab has a compact representation, i.e.,
\begin{align*}
 {\rm Lateral~slabs}\quad     &\left\{{\bf X}^{(1)}_j=\underline{\bf X}(:,j,:)={\bf  C}{\bf D}_j({\bf  B}){\bf  A}^T\right\}_{j=1}^J,\\
	 {\rm Frontal~slabs}\quad   		&\left\{{\bf X}^{(2)}_k=\underline{\bf X}(:,:,k)={\bf  A}{\bf D}_k({\bf  C}){\bf  B}^T\right\}_{k=1}^K,\\
		 {\rm Horizontal~slabs}\quad   		&\left\{{\bf X}^{(3)}_i=\underline{\bf X}(i,:,:)={\bf  B}{\bf D}_i({\bf  A}){\bf  C}^T\right\}_{i=1}^I,
\end{align*}
where ${\bf D}_r({\bf X}) = {\rm Diag}\left({\bf X}(r,:)\right)$.
Fig.~\ref{fig:tensor_1} gives a visual illustration of a three-way tensor $\underline{\bf X}$ and its slabs.

Unlike matrix factorizations, which are in general non-unique, the PARAFAC decomposition has (essentially) unique solution under quite mild conditions.
For example, Kruskal proved the following result for a real-valued low-rank tensor \cite{Kruskal1977}:
If
\begin{equation}\label{eq:Kruskal}
k_{\bf A}+k_{\bf B}+k_{\bf C}\geq 2R+2,
\end{equation}
then
$({\bf A},{\bf B},{\bf C})$ are unique up to a common column permutation and scaling, i.e.,
$\underline{\bf X}=\sum_{r=1}^R{\bf A}(:,r)\circ{\bf B}(:,r)\circ{\bf C}(:,r) = \sum_{r=1}^R\bar{\bf A}(:,r)\circ\bar{\bf B}(:,r)\circ\bar{\bf C}(:,r)$ $\Rightarrow$ $\bar{\bf A}={\bf A}{\bm \Pi}{\bm \Delta}_a$, $\bar{\bf B}={\bf B}{\bm \Pi}{\bm \Delta}_b$, $\bar{\bf C}={\bf C}{\bm \Pi}{\bm \Delta}_c$, where ${\bm \Pi}$ is a permutation matrix and ${\bm \Delta}_a$,  ${\bm \Delta}_b$,  ${\bm \Delta}_c$, are full-rank diagonal matrices such that $ {\bm \Delta}_a{\bm \Delta}_b{\bm \Delta}_c = {\bf I}$.
%This result has been generalized to the complex-valued case later; see \cite{sidiropoulos2000blind}.
If ${\bf A}$ is drawn from an absolutely continuous distribution over $\mathbb{R}^{I\times R}$, then $k_{\bf A}$ $=$ $\text{rank}({\bf A})$ $=$ $\text{min}(I,R)$ with probability one. It follows that if ${\bf A},{\bf B},{\bf C}$ are drawn this way, then the condition in \eqref{eq:Kruskal} can be simplified: if
\begin{equation}\label{eq:Kruskal_random}
\min\{I,R\}+\min\{J,R\}+\min\{K,R\}\geq 2R+2,
\end{equation}
then $({\bf A},{\bf B},{\bf C})$ are unique up to a common column permutation and scaling, with probability one.
Notice that under the condition in~\eqref{eq:Kruskal} or \eqref{eq:Kruskal_random}, the loading matrices ${\bf A}, {\bf B}, {\bf C}$ need not be tall. This is advantageous in challenging application scenarios, e.g., mixing system identification when the system is under-determined \cite{RonVorGerSid:2005,lee2013khatri}.
%To show the above, we have used the property that $k_{\bf X}=\min\{m,n\}$ for a matrix ${\bf X}\in\mathbb{R}^{m\times n}$ whose elements are drawn from some absolutely continuous distribution.

%From an estimation point of view, the three loading matrices are often obtained via the following optimization-based formulation:
{In practice, when modeling error and noise exist, it makes more sense to seek the best rank-$R$ approximation of a tensor rather than computing its exact rank factorization. To find such an approximation, the least squares criterion is commonly adopted}:
\begin{equation}\label{eq:LS-PARAFAC}
\min_{{\bf A},{\bf B},{\bf C}}~\left\|\underline{\bf X}- \sum_{r=1}^R{\bf A}(:,r)\circ{\bf B}(:,r)\circ{\bf C}(:,r)\right\|_F^2.
\end{equation}
{The above problem is nonconvex, and thus could be very difficult to solve}.
{In fact, recent research \cite{lim2014blind} showed that Problem~\eqref{eq:LS-PARAFAC} may even be `ill-posed', meaning that the best rank-$R$ approximation of a tensor may not even exist}.
{In practice, nevertheless, the formulation in \eqref{eq:LS-PARAFAC} allows one to devise computationally affordable (albeit generally suboptimal) algorithms, and some of these algorithms have proven successful in various applications.}
To deal with the optimization problem in \eqref{eq:LS-PARAFAC}, a popular way is to make use of the \emph{matrix unfoldings} of the tensor.
Specifically, by {vectorizing} each type of slabs and treating them as columns of a matrix, we obtain the three matrix unfoldings, namely,
$\underline{\bf X}^{(1)}=({\bf A}\odot{\bf C}){\bf B}^T$, $\underline{\bf X}^{(2)}=({\bf B}\odot{\bf A}){\bf C}^T$, and $\underline{\bf X}^{(3)}=({\bf C}\odot{\bf B}){\bf A}^T$,
where we have used the vectorization property of the Khatri-Rao product ${\rm vec}({\bf X}{\rm Diag}({\bf z}){\bf Y}^T)=({\bf Y}\odot{\bf X}){\rm vec}({\bf z})$.
Using the unfoldings, Problem~\ref{eq:LS-PARAFAC} can be tackled by cyclically solving the following three least squares problems:
\begin{subequations}
\begin{align}
          {\bf B}&:=\arg\min_{\bf B}~\|\underline{\bf X}^{(1)}-({\bf A}\odot{\bf C}){\bf B}^T\|_F^2\\
					{\bf C}&:=\arg\min_{\bf C}~\|\underline{\bf X}^{(2)}-({\bf B}\odot{\bf A}){\bf C}^T\|_F^2\\
					{\bf A}&:=\arg\min_{\bf A}~\|\underline{\bf X}^{(3)}-({\bf C}\odot{\bf B}){\bf A}^T\|_F^2.
\end{align}
\end{subequations}
The above updates yield the popular \emph{trilinear alternating least squares} (TALS) algorithm \cite{sidiropoulos2000blind,RonVorGerSid:2005}.

Although quite a lot of different PARAFAC algorithms exist, e.g., \cite{lee2013khatri,SOBIUM,Nion2008,Tichavsky2011,ACDC},
TALS (and its close relatives) has been the workhorse of low-rank tensor decomposition for decades for several reasons:
First, TALS can be easily implemented, since each iteration only involves relatively simple linear least squares subproblems. Second, it features monotone convergence of the cost function, without the need to tune (e.g., step-size) parameters to ensure this.
Third, it has the flexibility to incorporate constraints and regularization on the loading factors under its alternating optimization framework, with a reasonable complexity increase.
%Third, TALS can be scaled up to deal with big data problems by taking advantage of the sparsity of data.

\vspace*{-0.1in}
\section{A Closer Look At Motivating Examples}\label{sec:motivation}
%A wide range of applications, from classic array processing [XXX], wireless communication [XXX], analytic chemistry, to very recent topics such as social network mining [XXX], satisfy the PARAFAC model.
%In this work, we pay special attention to the scenario wherein some slabs of the PARAFAC model is highly corrupted.
%To motivate our work, several application examples are given as follows.

In many applications, some slabs of the collected tensor data are highly corrupted, for various reasons.
In this section, we take a closer look at some pertinent examples that we have encountered in rather different fields.
In all of them, {corrupted slabs} can throw off the analysis, producing inconsistent and hard to interpret PARAFAC models.

\subsection{Blind Speech Separation}
It has been shown that PARAFAC can be applied to blind speech separation (BSS) to identify the mixing system \cite{Nion2010,Reilly2005}.
As a quick review, the BSS signal model is
\begin{equation}\label{eq:instant_mix}
{\bf x}(t) = {\bf A}{\bf s}(t) + {\bf n}(t),~t=1,2,...
\end{equation}
where ${\bf x}(t)=[x_1(t),\ldots,x_I(t)]^T\in\mathbb{R}^{I}$ denotes the received signals by the $I$ sensors at time $t$,
${\bf A}\in\mathbb{R}^{I\times R}$ denotes the mixing system,
${\bf s}(t)=[s_1(t),\ldots,s_R(t)]^T\in\mathbb{R}^{R}$ denotes the $R$ speech sources (presumed to be uncorrelated),
and ${\bf n}(t)=[n_1(t),\ldots,n_I(t)]^T\in\mathbb{R}^{I}$ denotes zero-mean i.i.d. Gaussian noise with variance $\sigma^2$.
To connect this model to the PARAFAC model, we calculate the local covariance of the received signals within time frame $k$ by
\begin{equation*}
\begin{aligned}
\underline{\bf X}(:,:,k) &= \mathbb{E}\{{\bf x}(t){\bf x}^T(t)\} - \hat{\sigma}^2{\bf I}\\
&\approx {\bf A}\mathbb{E}\{{\bf s}(t){\bf s}^T(t)\}{\bf A}^T,\quad t\in [(k-1)L+1,kL],
\end{aligned}
\end{equation*}
where $\hat{\sigma}^2$ represents the estimated noise variance and $L$ denotes the time frame length.
By assuming that the sources are uncorrelated, we see that the local covariance of the sources in frame $k$, i.e., for $t\in [(k-1)L+1,kL]$,
\[\mathbb{E}\{{\bf s}(t){\bf s}^T(t)\}={\rm Diag}([\mathbb{E}|s_1(t)|^2,\ldots,\mathbb{E}|s_R(t)|^2]),\]
is a diagonal matrix.
Hence, if we let ${\bf C}(k,:)=[\mathbb{E}|s_1(t)|^2,\ldots,\mathbb{E}|s_R(t)|^2]$ for $t\in [(k-1)L+1,kL]$,
we see that $\underline{\bf X}(:,:,k)={\bf A}{\bf D}_k({\bf C}){\bf A}^T$ is a frontal slab of a three-way tensor (with ${\bf B}={\bf A}$),
and thus PARAFAC can be applied to $\underline{\bf X}$ to estimate the mixing system ${\bf A}$.
Using the estimated $\hat{\bf A}$, the individual source signals can be estimated. In the presence of reverberation, the mixing system model becomes convolutive (i.e., frequency-selective) instead of instantaneous. This is a more challenging scenario, which can again be tackled using PARAFAC in the frequency domain, see \cite{Nion2010,Reilly2005,FuMaHuaSid2015} and references therein.

%e.g., by $\hat{\bf s}(t)=\hat{\bf A}^\dag{\bf x}(t)$. In the presence of reverberation, the mixing system model becomes convolutive (i.e., frequency-selective) instead of instantaneous. This is a more challenging scenario, which can again be tackled using PARAFAC in the frequency domain, see \cite{Nion2010,Reilly2005}.

A more subtle difficulty is that some speech sources exhibit (strong) short-term cross correlations, even though they are approximately uncorrelated over the long run.
Consequently, the local covariances of the sources in some frames have significant off-diagonal elements, and the corresponding slabs deviate from the nominal model $\underline{\bf X}(:,:,k)={\bf A}{\bf D}_k({\bf C}){\bf A}^T$.
In such cases, directly applying standard PARAFAC algorithms may not yield satisfactory speech separation performance \cite{FuMaHuaSid2015}.

\subsection{Fluorescence Spectroscopy}
%In chemistry, fluorescence emission spectra measured at several excitation wavelengths for multiple samples forms a
%three-way low-rank tensor.
Fluorescence excitation-emission measurements (EEMs) are used in many
different fields such as skin analysis, fermentation monitoring,
environmental, food, and clinical analysis \cite{bro2011eemizer}.
A fluorescence sample is obtained by using a beam of light that excites the electrons in molecules of certain compounds and causes them to emit light;
the emission spectra are then measured at several excitation wavelengths.
A fluorescence EEM sample can be represented by
\[\underline{\bf X}(i,:,:)={\bf B}{\bf D}_i({\bf A}){\bf C}^T,\]
where ${\bf B}(:,r)$ for $r=1,\ldots,R$ corresponds to the spectral emission $r$,
${\bf C}(:,r)$ denotes the corresponding excitation values, and ${\bf A}(i,r)$ denotes the corresponding concentration (scaling) at sample $i$.
By measuring multiple samples, a PARAFAC model can be formed, and each sample is a slab.

Fluorescence data analysis has been recognized as a very successful example of applying PARAFAC algorithms to real-world data.
At the same time, it has also been noticed that anomalous EEM samples occur frequently
due to various reasons \cite{engelen2011detecting,hubert2012robust,bro2011eemizer,CEM:CEM1208}.
%\reminder{From Xiao to Rasmus: I still would like to put at least one example here, explaining why strange samples could happen. Could you give us an (very short) example?}

%For example, Rayleigh (and Raman) scattering creates highly irrelevant values that compromise the underlying low-rank tensor model.
%In practice, the scattering effect is usually detected and compensated by some specialized pre-processing algorithms automatically.
%However, at some slabs, the scattering effect can be too strong to be effectively removed or compensated, which leaves these slabs as outliers.

%, which admittedly degrades the effectiveness of applying the PARAFAC algorithm .
%For example, the Rayleigh (and Raman) scattering creates highly irrelevant values that compromise the underlying low-rank tensor structure, and makes the PARAFAC decomposition tend to fail \cite{bro2011eemizer};
%at some slabs, the scattering effect could be so strong to be removed or compensate, which leaves these slabs being outliers.
%We notice that scattering usually happens across all the samples, but at some samples it might be particularly severe.
%Hence, a fluorescence data set can be considered as a tensor with some badly contaminated slabs and some slightly corrupt but still informative slabs.
%A PARAFAC algorithm that can automatically select the latter and fit them with a PARAFAC model is therefore desired.

\subsection{Social Network Mining}
For some three-way social network data sets,
every slab $\underline{\bf X}(:,:,k)$ is a connected graph measured within time period $k$.
For example, in the ENRON e-mail data set \cite{diesner2005communication}, $\underline{\bf X}(i,j,k)$ denotes the `connection intensity' of person $i$ and person $j$ at time period $k$ (i.e., the number of e-mails sent by person $i$ to person $j$ within month $k$).
Another example is the Amazon purchase data. There, $\underline{\bf X}(i,j,k)$ represents the amount of product $j$ bought by person $i$ in week $k$.
%Hence, a slab $\underline{\bf X}(:,:,k)$ represents a connection graph that is measured within a certain period $k$.
For such data, each rank-one component of the PARAFAC model can be interpreted as the interaction pattern of a social group over time \cite{papalexakis2013k}.
To be specific, consider
\[\underline{\bf X}(:,:,k)\approx{\bf A}{\bf D}_k({\bf C}){\bf B}^T=\sum_{r=1}^R{\bf C}(k,r){\bf A}(:,r)({\bf B}(:,r))^T.\]
%Let ${\bf X}(i,j,k)$ denote the `connection intensity' of person $i$ and person $j$ at time period $k$ (e.g., the number of e-mails sent by person $i$ to person $j$within time period $k$) - it can also represent the purchase amount of product $j$ by person $i$ in week $k$, the number of posts by person $i$ on the facebook timeline (homepage) of person $j$, and so on.
Here, the nonzero elements in ${\bf A}(:,r)$ and ${\bf B}(:,r)$ create a clique (a subgraph) ${\bf A}(:,r){\bf B}^T(:,r)$, which can be interpreted as a social group,
{and $R$ corresponds to the number of social groups}.
Taking the ENRON e-mail data as an example, ${\bf A}(:,r){\bf B}^T(:,r)$ is a group, where the people corresponding to the non-zero elements of ${\bf A}(:,r)$
have similar e-mail sending patterns to those corresponding to the non-zero elements of ${\bf B}(:,r)$.
${\bf C}(k,r)$ is a time-varying parameter of this group, which means that the e-mail sending pattern of this group is a rank-one matrix factor whose intensity (e-mail volume) varies with time.

With this model, factoring the data box into its latent factors is equivalent to mining the underlying social groups, which finds applications in designing recommendation systems, analyzing ethic and cultural groups, and even detecting criminal organizations.
However, the social network data sets are in general not following a generative signal model, which means that several slabs may have large modeling errors.
As we will see later, some unexpected events (such as the ENRON crisis) might make the group e-mail patterns quite irregular during some period.
The slabs measured in these irregular time intervals might need to be identified and somehow down-weighted when the objective is to analyze the normal interaction patterns, or to detect those anomalies.

\vspace*{-0.1in}
\section{Problem Formulation}
\begin{figure}
	\centering
	\psfrag{xABC}{~}
	\psfrag{CORRUPT}{corrupted slabs}
		\includegraphics[width=8cm]{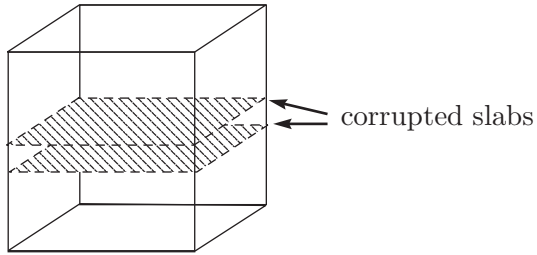}
	\caption{The corruption model: some horizontal slabs are outliers.}
	\label{fig:Corrupt_slab}
\end{figure}

%To address the problems that have been mentioned in the motivating examples,
%our objective is to propose an algorithm framework that can automatically fit a PARAFAC model using the `good' slabs;
%we wish our algorithm is not much more complicated than TALS, and is flexible to add regulizers and constraints.
Motivated by the examples in the previous section, we will focus on modeling, formulating, and solving the low-rank tensor decomposition problem in the presence of outlying slabs.
Our main goal is an easily implemented optimization framework;
practical considerations such as regularization, constraints, initialization and complexity will also be discussed.
For presentation simplicity, we will assume that corruption happens in some horizontal slabs throughout the development of the algorithm; see Fig.~\ref{fig:Corrupt_slab}.
Algorithms dealing with corrupt lateral or frontal slabs can be obtained by simply permuting the modes of the tensor, by virtue of symmetry.

To begin with, let us assume that some horizontal slabs have been corrupted by gross errors; i.e., we have
\begin{equation}
{\bf X}^{(3)}_i =\begin{cases}{\bf  B}{\bf  D}_i({\bf  A}){\bf  C}^T + {\bf O}_i,&\quad i\in {\cal N},\\ {\bf  B}{\bf  D}_i({\bf  A}){\bf  C}^T, &\quad i\in {\cal N}_c,\end{cases}
\label{eq:corrupt_model}
\end{equation}
where ${\cal N}\subset \{1,\ldots,I\}$ is the index set of the \emph{outlying slabs} and ${\cal N}_c=\{1,\ldots,I\}-{\cal N}$.
The gross error component ${\bf O}_i$ could be strong so that ${\bf X}^{(3)}_i$ is far from the nominal `clean signal model', i.e., ${\bf X}^{(3)}_i={\bf  B}{\bf  D}_i({\bf  A}){\bf  C}^T$.
Under the corruption model in \eqref{eq:corrupt_model}, our first observation here is that there may still be enough clean data to enable us to recover ${\bf B}$ and ${\bf C}$ intact. Thus, our idea begins with a formulation that guarantees the identifiability of ${\bf B}$ and ${\bf C}$ under some conditions.
%Notice that the information of ${\bf A}(i,:)$ for $i\in{\cal N}$ might be missing,
%while such ${\bf A}(i,:)$'s may be reconstructed by interpolation using some \emph{a priori} information - this part will be considered later in Section~\ref{sec:constraints}.

%an additional step of estimating ${\bf A}$ using the estimated ${\bf B}$ and ${\bf C}$ will also be discussed later.

%In the literature [XX,XXX], fitting a PARAFAC model with corrupted slabs have been considered.
%There, the intuition is to select some slabs following some heuristics at each iteration, and then perform plain PARAFAC multiple times.

We wish to fit the clean data slabs with a PARAFAC model.
In practice, ${\cal N}$ is usually unknown, but its cardinality may be small relative to $I$. Hence, we address this problem from a group-sparsity promoting viewpoint. We formulate the problem as
\begin{equation}\label{eq:L0Fit}
	         \min_{{\bf  A},{\bf  B},{\bf  C}}~\sum_{i=1}^I{\cal I}\left(\left\| \underline{{\bf  X}}^{(3)}(:,i) - ({\bf  C}\odot{\bf  B})({\bf  A}(i,:))^T \right\|_2\right),
\end{equation}
%\vspace{\baselineskip}
%\noindent
%\fbox{
%\parbox{.95\linewidth}{
%{\bf Criterion:}
%\begin{equation*}\label{eq:L0Fit}
%\begin{aligned}
	         %\min_{{\bf  A},{\bf  B},{\bf  C},{\bf Y}}\quad&\|{\bf Y}\|_{2,0}\\
       		%{\rm s.t.}\quad		&{\bf Y}(:,i) = \underline{{\bf  X}}^{(3)}(:,i) - ({\bf  C}\odot{\bf  B}){\bf  A}^T(i,:),\quad \forall i,
%\end{aligned}
%\end{equation*}}
%}
%\vspace{\baselineskip}
%
%\noindent
%where ${\bf Y}\in\mathbb{C}^{KJ\times I}$ is a slack variable，
%$\|{\bf Y}\|_{2,0}=\sum_{i=1}^I{\cal I}(\|{\bf y}_i\|_2)$ measures the column sparsity order of ${\bf Y}$, and
where ${\cal I}(x)$ is defined as
\[{\cal I}(x)=\begin{cases}1,&~x\neq 0\\0,&~x=0.\end{cases}.\]
The criterion tends to make $\underline{{\bf  X}}^{(3)}(:,i) - ({\bf  C}\odot{\bf  B})({\bf  A}(i,:))^T={\bf 0}$ for as many $i$'s as possible.
Intuitively, if there are enough clean slabs to identify the underlying nominal PARAFAC model, solving the above optimization problem should identify ${\bf B}$ and ${\bf C}$.
The following result confirms this intuition.
\begin{Claim}\label{thm:L0PARAFAC}
Assume that the elements of ${\bf A}$ are drawn from an absolutely continuous distribution over $\mathbb{R}^{I\times R}$, and likewise ${\bf B}$ and ${\bf C}$ are drawn from absolutely continuous distributions over $\mathbb{R}^{J\times R}$ and $\mathbb{R}^{K\times R}$, respectively. Define
\[c := 2R+2-\min\{J,R\} - \min\{K,R\},\]
and suppose that $c \leq \min \{|{\cal N}_c|,R\}$, and
\begin{equation}
    |{\cal N}_c|\geq \frac{I+c}{2}.
\end{equation}
Then, with probability one, the optimal ${\bf B}^\star$, ${\bf C}^\star$, and ${\bf A}^\star({\cal N}_c,:)$ that solve Problem~\eqref{eq:L0Fit} are ${\bf B}$, ${\bf C}$, and ${\bf A}({\cal N}_c,:)$ with a common column permutation and scaling; i.e., ${\bf A}^\star({\cal N}_c,:)={\bf A}({\cal N}_c,:){\bm \Pi}{\bm \Delta}_a$, ${\bf B}^\star={\bf B}{\bm \Pi}{\bm \Delta}_b$, ${\bf C}^\star={\bf C}{\bm \Pi}{\bm \Delta}_c$, where ${\bm \Pi}$ is a permutation matrix and ${\bm \Delta}_a$,  ${\bm \Delta}_b$,  ${\bm \Delta}_c$, are full-rank diagonal matrices such that $ {\bm \Delta}_a{\bm \Delta}_b{\bm \Delta}_c = {\bf I}$.
\end{Claim}

The proof of Claim~\ref{thm:L0PARAFAC} can be found in Appendix~\ref{app:identifiability}.
Claim~\ref{thm:L0PARAFAC} helps us understand the fundamental limitation of the proposed criterion in \eqref{eq:L0Fit}:
Under the signal model in \eqref{eq:corrupt_model}, if about one half of the horizontal slabs follow the clean signal model, we can still correctly identify the two loading factors ${\bf B}$ and ${\bf C}$ (and at least part of ${\bf A}$).
Solving Problem~\eqref{eq:L0Fit} is very challenging though - both PARAFAC decomposition and group-sparsity maximization (cardinality minimization) are {nonconvex} problems on their own, so \eqref{eq:L0Fit} is compounding two already challenging problems.
%Plus, although Problem~\eqref{eq:L0Fit} provides clean-cut identifiability insight, such type of criterion is not robust against modeling error and noise.
In the next section, a more practical optimization surrogate will be employed to approximate Problem~\eqref{eq:L0Fit}, and a simple alternating optimization algorithm will be presented to tackle this surrogate optimization problem.

%We should mention that if the outliers are rather random, it can be easily shown that ${\bf B}$ and ${\bf C}$ can be identified almost surely as long as there is an identifiable sub-cube of the whole data under our signal model (in words, even if $99\%$ percent of slabs are random outliers, we can still identify ${\bf B}$ and ${\bf C}$ by solving Problem~\eqref{eq:L0Fit}).
\section{Basic Algorithmic Framework}
To approximate Problem~\eqref{eq:L0Fit}, we propose to employ the smoothed $\ell_p$ quasi-norm as our working objective;
i.e., by replacing $\sum_{i=1}^I{\cal I}(x_i)$ by $\sum_{i=1}^I(x_i^2+\epsilon)^{p/2}$,
we deal with the following surrogate:
\begin{equation}\label{eq:LpFit}
	         \min_{{\bf  A},{\bf  B},{\bf  C}}~\sum_{i=1}^I\left(\left\| \underline{\bf  X}^{(3)} (:,i)- ({\bf  C}\odot{\bf  B})({\bf  A}(i,:))^T \right\|_2^2+\epsilon\right)^{p/2},
\end{equation}
where $0<p\leq 1$ and $\epsilon>0$.
The idea comes from compressive sensing, where
the quasi $\ell_0$ norm is often approximated by the quasi $\ell_p$ norm or $\ell_1$ norm, since the latter two are computationally tractable and often yield practically good results \cite{chartrand2008restricted,Rao1999,Chartrand2008,chartrand2007nonconvex}.
Here, $\epsilon$ is a small smoothing parameter to keep the cost function in its continuously differentiable domain.

%
%\begin{figure}[!h]
	%\centering
		%\includegraphics[width=8cm]{norms.eps}
	%\caption{Using $\ell_p$-norm $(0<p\leq 1)$ to approximate $\ell_0$-norm.}
	%\label{fig:norms}
%\end{figure}

The cost function in \eqref{eq:LpFit} can be manipulated according to the following lemma:
%\begin{Lemma}\label{lem:conjugate}
    %Assume $0<p\leq 2$ and
		%\[\phi_p(w_i) = \frac{2-p}{2}\left(\frac{2}{p}w_i \right)^{\frac{p}{p-2}} -\epsilon w_i.\]
		%Then, we have
		%\begin{align*}
		  %&\left(\left\| \underline{{\bf  X}}^{(3)}(:,i) - ({\bf  C}\odot{\bf  B})^T{\bf  A}(i,:) \right\|_2^2+\epsilon\right)^{p/2} \\
		 %=&\min_{w_i\geq 0}~{w}_i\left\| \underline{{\bf  X}}^{(3)}(:,i) - ({\bf  C}\odot{\bf  B}){\bf  A}^T(i,:) \right\|_2^2+\phi_p(w_i)
		%\end{align*}
		%and the unique minimizer is
		%\[w_i^\star = \frac{p}{2}\left(\left\| \underline{{\bf  X}}(:,i)^{(3)} - ({\bf  C}\odot{\bf  B}){\bf  A}^T(i,:) \right\|_2^2+\epsilon\right)^{\frac{p-2}{2}}.\]
%\end{Lemma}

\begin{Lemma}\label{lem:conjugate}
    Assume $0<p<2$, $\epsilon\geq 0$, and
		$\phi_p(w) := \frac{2-p}{2}\left(\frac{2}{p}w \right)^{\frac{p}{p-2}} +\epsilon w$.
		Then, we have
		\begin{align*}
		  &\left(x^2+\epsilon\right)^{p/2} = \min_{w\geq 0}~w x^2+\phi_p(w),
		\end{align*}
		and the unique minimizer is
		\begin{equation}
    w_{\rm opt} = \frac{p}{2}\left(x^2+\epsilon\right)^{\frac{p-2}{2}}.
		\label{eq:w_opt}
		\end{equation}
\end{Lemma}

\emph{Proof}:
     First, it can be seen that $\phi_p(w)$ is strictly convex on its domain (i.e., the interior of $w\geq 0$), since its second order derivative is positive when $w$ is positive, i.e.,
		 \[\nabla^2\phi_p(w)= -\frac{4}{p(p-2)}\left(\frac{2}{p}w \right)^{\frac{4-p}{p-2}}>0.\]
		 %where $\phi_p''(w)$ denotes the second order derivative of $\phi_p(w)$,
		 %and $\phi_p(w)=\infty$ for $w=0$.
		 Therefore,
		 \begin{equation}
		 \min_{w\geq 0}~{w}x^2+\phi_p(w)
		 \label{eq:g_wm}
		\end{equation}
		 admits a unique optimal solution
		 $w_{\rm opt} = (p/2)(x^2+\epsilon)^{(p-2)/2}$, which can be obtained by simply checking the first order optimality condition.
		 Substituting $w_{\rm opt}$ back into the cost of \eqref{eq:g_wm}, the minimum cost is $(x^2+\epsilon)^{p/2}$.
\hfill $\square$

By Lemma~\ref{lem:conjugate}, Problem~\eqref{eq:LpFit} can be re-expressed as the following problem:
\begin{equation}
\begin{aligned}
	         \min_{\begin{subarray}{c}{\bf  A},{\bf  B},{\bf  C},\\ \{w_i\geq 0\}\end{subarray}}~\sum_{i=1}^I w_i\left\| \underline{{\bf  X}}^{(3)}(:,i) - ({\bf  C}\odot{\bf  B}) \left({\bf  A}(i,:)\right)^T \right\|_2^2+\sum_{i=1}^I\phi_p(w_i).
\end{aligned}
\label{eq:w_PARAFAC}
\end{equation}
The structure of Problem~\eqref{eq:w_PARAFAC} is nice: it allows us to optimize its cost with respect to (w.r.t.) the four blocks ${\bf  A},{\bf  B},{\bf  C}$, and $\{w_i\}_{i=1}^I$
in an alternating optimization fashion, fixing three blocks and updating one each time\footnote{A similar auxiliary variable-based technique for splitting {\em convex} $\ell_p$ norms ($1\leq p<2$) has appeared in \cite{geman1992constrained,idier2001convex}.
%As mentioned in \cite{geman1992constrained}, the potential benefits of splitting are unclear for convex problems such as those considered in \cite{geman1992constrained,idier2001convex}.
Lemma~1 can be considered as a nonconvex extension of the prior works in \cite{geman1992constrained,idier2001convex}.}.
As we will show next, each conditional optimization problem has a closed-form solution.

%Interestingly, the authors commented that casting convex $\ell_p$
%such a technique is much more helpful in the nonconvex case.

%\begin{Remark}
%Lemma~\ref{lem:conjugate} is not only useful for the ensuing algorithmic development, it can also enable similar progress in developing alternating $\ell_p$-optimization algorithms for other, potentially very different models.
%\end{Remark}

First, the problem w.r.t. ${\bf  A}$ is separable w.r.t. $i$. For each $i$, the problem w.r.t. ${\bf A}(i,:)$ is a simple least squares problem.
Hence, the subproblem w.r.t. ${\bf A}$ admits the following closed-form solution:
\begin{equation*}\label{eq:A_sln}
	      {\bf  A}=  \left(  ({\bf  C}\odot{\bf  B})^\dag  \underline{{\bf  X}}^{(3)} \right)^T,
\end{equation*}
which is the same as that in the plain TALS \cite{sidiropoulos2000blind,RonVorGerSid:2005}.
Notice that in practice, we compute ${\bf A}$ by the following expression:
\begin{equation*}\label{eq:A_prac}
{\bf A}^T = ({\bf C}^T{\bf C}\circledast{\bf B}^T{\bf B})^{-1}({\bf C}\odot{\bf B})^T\underline{\bf X}^{(3)}.
\end{equation*}
In practice, the matrix inversion part and $({\bf C}\odot{\bf B})^T\underline{\bf X}^{(3)}$ should be computed separately.
The reasons are as follows.
First, the inversion part, i.e., $({\bf C}^T{\bf C}\circledast{\bf B}^T{\bf B})^{-1}$, is usually the inverse of a small ($R$-by-$R$) matrix.
Second, the multiplication of a Khatri-Rao structured matrix and an unfolded tensor is a computationally expensive operation if $I,J,K$ are large (specifically, this single step costs $2RIJK$ flops),
but fast algorithms are available when $\underline{\bf X}$ is sparse \cite{BaderKolda2007,kolda2008scalable,TensorToolbox}, \cite{papalexakis2013large}, \cite{RavSidSmiKar:Asilo2014}.

%The reasons lie in computation efficiency and memory saving.
%Specifically, for big data problems, the operation with the form of a product of a Khatri-Rao product term and a big matrix (e.g.,
%$({\bf C}\odot{\bf B})^T\underline{\bf X}^{(3)}$) has very efficient solvers [XXX,XXX,XXX].
%Also, using [Kolda], one can finish this step without metricizing the tensor and save three big matrix unfoldings, which may save a large amount of memory.

To update ${\bf  B}$, we consider using the lateral slabs $\{{\bf  C}{\bf  D}_j({\bf  B}){\bf  A}^T\}_{j=1}^J$.
From Problem~\eqref{eq:w_PARAFAC}, it can be readily seen that the $i$th column of $\{{\bf  C}{\bf  D}_j({\bf  B}){\bf  A}^T\}$ is scaled by $\sqrt{w_i}$.
Thus, the subproblem w.r.t. ${\bf B}$ can be written as
\[       \min_{{\bf  B}}~\sum_{j=1}^J\left\|{\bf  X}_{j}^{(1)}{\bf  W}-{\bf  C}{\bf  D}_j({\bf  B}){\bf  A}^T{\bf  W}\right\|_F^2,\]
where $ {\bf  W}={\rm Diag}(\sqrt{w_1},\ldots,\sqrt{w_I})$,
or, in the following more compact form,
\[    \min_{{\bf  B}}~\left\|({\bf  W}\otimes{\bf  I})\underline{\bf  X}^{(1)}-\left(({\bf  W}{\bf  A})\odot{\bf  C}\right){\bf  B}^T\right\|_F^2.\]
The above is still a least squares problem.
Therefore, the solution is simply
\begin{equation*}\label{eq:B_sln}
 {\bf  B}=\left(\left(({\bf  W}{\bf  A})\odot{\bf  C}\right)^\dag({\bf  W}\otimes{\bf  I})\underline{\bf  X}^{(1)}\right)^T.
\end{equation*}
In practice, the above solution can be written as follows:
\begin{subequations}
\begin{align}
{\bf  B}^T&=\left({\bf  W}{\bf  A}\odot{\bf  C}\right)^\dag({\bf  W}\otimes{\bf  I})\underline{\bf  X}^{(1)},\\
           &=\left(  ({\bf  W}{\bf  A}\odot{\bf  C})^T({\bf  W}{\bf  A}\odot{\bf  C})\right)^{-1}({\bf  W}{\bf  A}\odot{\bf  C})^T({\bf  W}\otimes{\bf  I})\underline{\bf  X}^{(1)} \label{eq:kh_prodct}\\
					 &= \left(  {\bf  A}^T{\bf  W}^2{\bf  A}\circledast{\bf  C}^T{\bf  C} \right)^{-1}({\bf  W}^2{\bf  A}\odot{\bf  C})^T\underline{\bf  X}^{(1)}, \label{eq:khkr_prodct}
\end{align}
\end{subequations}
where we have used the property
\[({\bf U}_1\odot{\bf V}_1)^T({\bf U}_2\odot{\bf V}_2)={\bf U}_1^T{\bf U}_2\circledast{\bf V}_1^T{\bf V}_2\]
to obtain \eqref{eq:kh_prodct}, and
\[({\bf U}_1\otimes{\bf V}_1)^T({\bf U}_2\odot{\bf V}_2)={\bf U}_1^T{\bf U}_2\odot{\bf V}_1^T{\bf V}_2\]
to reach \eqref{eq:khkr_prodct}.
Putting ${\bf B}$ in the form of \eqref{eq:khkr_prodct} is important.
The reason is twofold:
First, one does not have to actually compute and save $({\bf  W}\otimes{\bf  I})\underline{\bf  X}^{(1)}$ since saving $IJK$ elements after each iteration is cumbersome when $I,J,K$ are large (e.g., for $I=J=K=100$, a million variables have to be saved in each iteration).
Second, the efficient solvers for computing the product of a Khatri-Rao structured matrix and an unfolded tensor can be directly applied to $({\bf  W}^2{\bf  A}\odot{\bf  C})^T\underline{\bf  X}^{(1)}$.

To update ${\bf  C}$, the rationale follows that of updating ${\bf B}$.
Specifically, as the $i$th row of each frontal slab is scaled by $\sqrt{w_i}$,  we have can express the conditional problem w.r.t. ${\bf C}$ as
	      \[         \min_{\bf  C}~\sum_{k=1}^K\left\|{\bf  W}{\bf  X}_{k}^{(2)}-{\bf  W}{\bf  A}{\bf  D}_k({\bf  C}){\bf  B}^T\right\|_F^2,                 \]
and the solution is also in closed form:
\begin{equation*}\label{eq:C_sln}
    {\bf  C}=\left(\left({\bf  B}\odot{\bf  W}{\bf  A}\right)^\dag ({\bf  I}\otimes{\bf  W})\underline{\bf  X}^{(2)}\right)^T.
\end{equation*}
Similar to the ${\bf B}$ case, we can express ${\bf  C}^T$ as
\begin{equation*}
{\bf  C}^T = \left({\bf B}^T{\bf B} \circledast {\bf A}^T{\bf W}^2{\bf A}\right)^{-1}({\bf B}\odot{\bf W}^2{\bf A})^T\underline{\bf  X}^{(2)},
\end{equation*}
and thus $\left({\bf B}^T{\bf B} \circledast {\bf A}^T{\bf W}^2{\bf A}\right)^{-1}$ and $({\bf B}\odot{\bf W}^2{\bf A})^T\underline{\bf  X}^{(2)}$ can be computed separately, if necessary in practice.

The update w.r.t. $\{w_i\}_{i=1}^I$ follows Lemma~\ref{lem:conjugate}, i.e.,
\begin{equation*}\label{eq:w_sln}
				 w_i:=\frac{p}{2}\left(\left\| \underline{{\bf  X}}^{(3)} - ({\bf  C}\odot{\bf  B}){\bf  A}^T(i,:) \right\|_2^2+\epsilon\right)^{\frac{p-2}{2}},\quad \forall i.
\end{equation*}

Given these conditional updates,
%several different types of updates can be implemented, each having different theoretical and practical properties.
%We summarize them as in the following:
%\subsubsection{Alternating Optimization}
a simple strategy is to cyclically update ${\bf A}$, ${\bf B}$, ${\bf C}$ and $\{w_1,\ldots,w_I\}$.
The algorithm is summarized in Algorithm~\ref{Algo:LpPARAFAC}; we will henceforth refer to it as \emph{Iteratively Reweighted Alternating Least Squares} (IRALS),
since $w_1,\ldots,w_I$ can be interpreted as weights applied to the frontal slabs.
{From an algorithmic structure viewpoint, IRALS can be considered as an extension of the iteratively reweighted least squares (IRLS) algorithm \cite{Chartrand2008} to tensor factorization}.
{Since each partial minimization does not increase the value of the cost function and the function is lower bounded by zero}, IRALS \emph{guarantees the convergence of the cost function of Problem~\eqref{eq:w_PARAFAC}}.
%This update is practically simple and efficient.

\begin{Remark}\label{rmk:convergence}

One may notice that we have not characterized the convergence of the solution sequence produced by IRALS yet.
By some existing theories of alternating optimization,
a stationary point for TALS and IRALS may be attained if the conditional objective function of every block is strictly convex and is continuously differentiable on the interior of the feasible set \emph{throughout all iterations} \cite[Proposition 2.7.1]{bertsekas1999nonlinear}.
In our context, this requires ${\rm rank}({\bf B}\odot{\bf A})= {\rm rank}({\bf C}\odot{\bf B}) ={\rm rank}({\bf A}\odot{\bf C})=R$ \emph{throughout all iterations}, which is hard to check \cite{li2013some}.
Nevertheless, convergence to a stationary point of Problem~\eqref{eq:LpFit} can be shown by employing some variants of alternating optimization, e.g., \emph{maximum block improvement} (MBI) \cite{MBI}.
%MBI compares the decreased objective value induced by each block at each iteration, and only updates the block that leads to the lowest objective value.
In this work, we adopt cyclic alternating optimization instead of MBI, for implementation simplicity and speed.
Also, in practice, we are often interested in PARAFAC with regularization on the loading factors;
in such cases, convergence to a stationary point of alternating optimization is usually not a problem any more \cite{li2013some} - see the next section for details.
\end{Remark}

\begin{Remark}
Until now, we have been dealing with the problem of interest (i.e., Problem~\eqref{eq:LpFit}) \emph{indirectly}.
It is interesting to consider the relationship between the solutions of our working problem, i.e., Problem~\eqref{eq:w_PARAFAC}, and Problem~\eqref{eq:LpFit}.
It can be shown that
\begin{Claim}\label{prop:convergence}
Assume that $({\bf  A}^\star,{\bf  B}^\star,{\bf  C}^\star,\{w_i^\star\}_{i=1}^I)$ is a stationary point of Problem~\eqref{eq:w_PARAFAC}.
Then, $({\bf  A}^\star,{\bf  B}^\star,{\bf  C}^\star)$ is also a stationary point of Problem~\eqref{eq:LpFit}.
\end{Claim}
The proof of Claim~\ref{prop:convergence} can be found in Appendix~\ref{app:claim2}.
%closely follows that of \cite[Proposition~2]{fu2015danser};
The key step is to invoke the uniqueness of the subproblem w.r.t. $\{w_i\}
$ following Lemma~\ref{lem:conjugate} and marginalize it.
By this claim, we see that dealing with Problem~\eqref{eq:w_PARAFAC} can yield a stationary point of Problem~\eqref{eq:LpFit},
whenever a limit point is reached.
\end{Remark}

\begin{algorithm}[t]
%\SetAlgoNoLine
\SetKwInOut{Input}{input}\SetKwInOut{Output}{output}
\SetKwRepeat{Repeat}{repeat}{until}

\Input{$\underline{\bf X}$; , ${\bf B}_0$, ${\bf C}_0$ (initialization); and $p\in(0,1]$.}

${\bf B}={\bf B}_0$;

${\bf C}={\bf C}_0$;

${\bf W}={\bf I}$;

\Repeat{some stopping criterion is satisfied}{

${\bf  A} :=  \left(  ({\bf C}^T{\bf C}\circledast{\bf B}^T{\bf B})^{-1}({\bf C}\odot{\bf B})^T\underline{\bf X}^{(3)} \right)^T$

${\bf  B} := \left(\left(  {\bf  A}^T{\bf  W}^2{\bf  A}\circledast{\bf  C}^T{\bf  C} \right)^{-1}({\bf  W}^2{\bf  A}\odot{\bf  C})^T\underline{\bf  X}^{(1)}\right)^T$;

${\bf  C} :=\left( \left({\bf B}^T{\bf B} \circledast {\bf A}^T{\bf W}^2{\bf A}\right)^{-1}({\bf B}\odot{\bf W}^2{\bf A})^T\underline{\bf  X}^{(2)}\right)^T$;

$w_i:=\frac{p}{2}\left(\left\| \underline{{\bf  X}}^{(3)}(:,i) - ({\bf  C}\odot{\bf  B}){\bf  A}^T (i,:)\right\|_2^2+\epsilon\right)^{\frac{p-2}{2}},\quad \forall i$;

${\bf W}^2:={\rm Diag}({w}_1,\ldots,{w}_I)$;

}

\Output{${\bf B}$, ${\bf C}$.}

\caption{IRALS}\label{Algo:LpPARAFAC}
\end{algorithm}

\vspace*{-0.1in}
\section{Extension: Constrained and Regularized Robust Tensor Factorization}\label{sec:constraints}
In this section, we consider practical extensions of IRALS, namely,
constrained and regularized optimization.

\subsection{Adding Constraints and Regularization}

In data analytics, constrained and regularized low-rank tensor factorization often makes a lot of sense,
since combining different types of \emph{a priori} information may help find interpretable factors when modeling error and noise exist.
Hence, there are many cases in which we are interested in solving the following problem:
\begin{equation}\label{eq:reg_cons_LpFit_0}
\begin{aligned}
	         \min_{{\bf  A},{\bf  B},{\bf  C}}&\quad\frac{1}{2}\sum_{i=1}^I\left(\left\| \underline{{\bf  X}}^{(3)}(:,i) - ({\bf  C}\odot{\bf  B}){\bf  A}^T(i,:) \right\|_2^2+\epsilon\right)^{p/2}\\
					&\quad\quad\quad\quad\quad+\lambda_a f({\bf A}) +\lambda_b g({\bf B}) + \lambda_c h({\bf C}),\\
					  {\rm s.t.}&\quad {\bf A}\in{\cal A}~,{\bf B}\in{\cal B},~{\bf C}\in{\cal C},
\end{aligned}
\end{equation}
where $\lambda_a$, $\lambda_b$ and $\lambda_c$ are nonnegative regularization parameters,
$f({\bf A})$, $g({\bf B})$ and $g({\bf C})$ are appropriate regularization functions,
and ${\cal A}$, ${\cal B}$ and ${\cal C}$ represent (hard) constraints on the loading factors.

In many cases, the constraints of interest include nonnegativity of the loadings, stemming from physical, chemical, or modeling considerations - e.g., concentrations, spectra, and e-mail counts are all nonnegative, and nonnegativity of the latent factors is important in social network mining \cite{papalexakis2013k} and in fluorescence spectroscopy \cite{bro2011eemizer}. More general `box' constraints of type $a_l\leq {\bf A}(i,r)\leq a_h$ may also be appropriate, e.g., when we also have prior knowledge on the maximum possible concentration.

Soft constraints may also be of interest, and these can be represented using appropriate regularization terms. If we know that the columns of ${\bf B}$ should be smooth, for example, we can employ the regularization ${g}({\bf B})=\|{\bf T}{\bf B}\|_F^2$, where \cite{CVX_BOOK}
\begin{equation}\label{eq:smooth_T}
{\bf T}=\begin{bmatrix} 1& -2 & 1     & 0    &\cdots &\cdots & \cdots \\
                          0& 1  &-2     & 1    & 0     &\cdots & \cdots\\
													\vdots&\vdots&\vdots &\vdots& \vdots& \vdots &\vdots\\
													\cdots&\cdots&\cdots &0      & 1     & -2    &1
													\end{bmatrix}.
													\end{equation}											
If we know that the loading factors are sparse, we can use $\|\cdot\|_1$ or any other sparsity-promoting function for regularization. As alluded to in Remark~\ref{rmk:convergence}, adding regularization often brings side-benefits in terms of accelerating convergence, avoiding swamps, and attaining a stationary point \cite{navascaswamp,li2013some}.
For example, by adding the minimum-norm regularization $\|{\bf A}\|_F^2$, $\|{\bf B}\|_F^2$, and $\|{\bf C}\|_F^2$, it can be easily seen that each block always has a unique solution, and thus a stationary point of Problem~\eqref{eq:reg_cons_LpFit_0} can be attained by alternating optimization, whenever a limit point exists.												 
Given the overall objective \eqref{eq:reg_cons_LpFit_0}, and by Lemma~\ref{lem:conjugate},
we may consider the equivalent reformulation
\begin{equation*}\label{eq:reg_cons_LpFit}
\begin{aligned}
	         \min_{{\bf  A},{\bf  B},{\bf  C},\{w_i\}}&\quad\sum_{i=1}^I\frac{w_i}{2}\left\| \underline{{\bf  X}}^{(3)}(:,i) - ({\bf  C}\odot{\bf  B}){\bf  A}^T(i,:) \right\|_2^2\\
					&+\frac{w_i}{2}\sum_{i=1}^I\phi_p(w_i)+\lambda_a f({\bf A}) +\lambda_b g({\bf B}) + \lambda_c h({\bf C}),\\
					  {\rm s.t.}&\quad {\bf A}\in{\cal A},~{\bf B}\in{\cal B},~{\bf C}\in{\cal C},\\
						          &\quad w_i\geq 0, \quad i=1,\ldots,I.
\end{aligned}
\end{equation*}
The subproblem w.r.t. $\{w_i\}$ admits the same solution as before.
In addition, the subproblems w.r.t. the loading factors are constrained and regularized least squares problems.
To describe our treatment, we begin with the subproblem w.r.t. ${\bf B}$:
\begin{equation*}
\begin{aligned}
      \min_{\bf B}&\quad\frac{1}{2}\left\|({\bf  W}\otimes{\bf  I})\underline{\bf  X}^{(1)} -\left(({\bf  W}{\bf  A})\odot{\bf  C}\right){\bf B}^T\right\|_F^2 + \lambda g({\bf B})\\
			{\rm s.t.}&\quad{\bf B}\in{\cal B}.
\end{aligned}
\end{equation*}
To handle this problem,
we propose the following \emph{alternating direction method of multipliers} (ADMM) \cite{Boyd11} based approach.
We first rewrite the problem as
\begin{equation}\label{eq:B-ADMM}
\begin{aligned}
      \min_{{\bf B},~{\bf B}_1,~{\bf B}_2}&\frac{1}{2}\quad\left\|({\bf  W}\otimes{\bf  I})\underline{\bf  X}^{(1)} -\left(({\bf  W}{\bf  A})\odot{\bf  C}\right){\bf B}_1^T\right\|_F^2\\
			 &\quad\quad\quad\quad+ \lambda g({\bf B}_2) + {\bf 1}_{\cal B}({\bf B})\\
			{\rm s.t.}&\quad{\bf B}={\bf B}_1\\
								&\quad{\bf B}={\bf B}_2.
\end{aligned}
\end{equation}
where ${\bf 1}_{\cal X}({\bf X})$ is $0$ for ${\bf X} \in {\cal X}$ and $\infty$ otherwise.
{ADMM solves the following augmented Lagrangian dual of Problem~\eqref{eq:B-ADMM} \cite[Chapter 3]{Boyd11}:}
\begin{equation}\label{eq:B-ADMM-dual}
\begin{aligned}
     {\max_{{\bf U}_1,~{\bf U}_2}} {\min_{{\bf B},~{\bf B}_1,~{\bf B}_2}}&{\frac{1}{2}\quad\left\|({\bf  W}\otimes{\bf  I})\underline{\bf  X}^{(1)} -\left(({\bf  W}{\bf  A})\odot{\bf  C}\right){\bf B}_1^T\right\|_F^2}\\
			 &\quad\quad\quad\quad{+ \lambda_b g({\bf B}_2) + {\bf 1}_{\cal B}({\bf B})}\\
			 &\quad\quad\quad\quad {+\frac{\rho}{2}\|{\bf B}-{\bf B}_1+ {\bf U}_1\|_F^2 }\\
			 &\quad\quad\quad\quad {+ \frac{\rho}{2}\|{\bf B}-{\bf B}_2 + {\bf U}_2\|_F^2},
\end{aligned}
\end{equation}
where ${\bf U}_1$ and ${\bf U}_2$ are the dual variables, and $\rho>0$ is the stepsize parameter that is pre-specified.
The standard ADMM updates for Problem~\eqref{eq:B-ADMM-dual} are as follows {\cite[Chapter 3]{Boyd11}}:
\begin{subequations}
\begin{align}
{\bf B}_1 &:= \arg\min_{{\bf B}_1} \quad\frac{1}{2}\left\|({\bf  W}\otimes{\bf  I})\underline{\bf  X}^{(1)} -\left(({\bf  W}{\bf  A})\odot{\bf  C}\right){\bf B}_1^T\right\|_F^2\nonumber\\
& \quad\quad +\frac{\rho}{2}\|{\bf B}-{\bf B}_1+{\bf U}_1\|_F^2 \label{eq:sub_B1}\\
{\bf B}_2 &:= \arg\min_{{\bf B}_2} \quad \lambda_b g({\bf B}_2) + \frac{\rho}{2}\|{\bf B}-{\bf B}_2 + {\bf U}_2\|_F^2 \label{eq:B_2_update}\\
{\bf B}& := \arg\min_{{\bf B}}\quad \frac{\rho}{2}\|{\bf B}-{\bf B}_2 + {\bf U}_2\|_F^2 \nonumber \\
& \quad\quad+\frac{\rho}{2}\|{\bf B}-{\bf B}_1+{\bf U}_1\|_F^2 + {\bf 1}_{\cal B}({\bf B}), \label{eq:sub_B}\\
{\bf U}_1& := {\bf U}_1 + {\bf B}-{\bf B}_1,\\
{\bf U}_2& := {\bf U}_2 + {\bf B}-{\bf B}_2,
\end{align}
\end{subequations}

The proposed variable-splitting strategy brings several advantages.
First, the problem w.r.t. ${\bf B}_1$ (i.e., Problem~\eqref{eq:sub_B1}) is a least squares problem, whose solution is
\[{\bf B}_1^T:=\left(  {\bf  A}^T{\bf  W}^2{\bf  A}\circledast{\bf  C}^T{\bf  C} + \rho{\bf I}\right)^{-1}\left(({\bf  W}^2{\bf  A}\odot{\bf  C})^T\underline{\bf  X}^{(1)}+{\bf M}\right),\]
where ${\bf M}=\rho({\bf B}+{\bf U}_1)$.
We see that the structure of $({\bf  W}^2{\bf  A}\odot{\bf  C})^T\underline{\bf  X}^{(1)}$ has been preserved, and thus efficient solvers for this matrix multiplication problem can be applied when the tensor is large and sparse \cite{BaderKolda2007,kolda2008scalable,TensorToolbox}, \cite{papalexakis2013large}, \cite{RavSidSmiKar:Asilo2014}.
Second, the ${\bf B}_2$ update is a proximal operator, which can be put in simple closed-form for many $g(\cdot)$'s.
Let us consider $g({\bf B}_2)=\frac{1}{2}\|{\bf T}{\bf B}_2\|_F^2$ as an example, which is often used for promoting smooth ${\bf B}$. The ${\bf B}_2$ update is then simply
\[  {\bf B}_2 := (\lambda_b{\bf T}^T{\bf T}+\rho{\bf I})^{-1}({\bf B}+{\bf U}_2).  \]
Note that when ${\bf T}={\bf I}$, this further reduces to ${\bf B}_2 =\frac{1}{\lambda_b + \rho}({\bf B}+{\bf U}_2)$ - which is useful to control the scaling of ${\bf B}$.
Also, if one wants to promote sparsity in ${\bf B}$, several convex and nonconvex $g(\cdot)$'s that enable closed-form solution of Problem~\eqref{eq:B_2_update} can be employed; see \cite{chartrand2014shrinkage}.
Third, the ${\bf B}$ update (Problem~\eqref{eq:sub_B}) also has a simple form:
\[{\bf B}:=\left(\frac{1}{2}({\bf B}_2 - {\bf U}_2+{\bf B}_1 - {\bf U}_1)\right)_{\cal B},\]
where $(\cdot)_{\cal B}$ is a projector to the set ${\cal B}$.
For many constraints ${\cal B}$, this projection step is fairly simple. For example,
if ${\cal B}=\mathbb{R}_{+}$, the projection is
\[{\bf B}:=\left(\frac{1}{2}({\bf B}_2 - {\bf U}_2+{\bf B}_1 - {\bf U}_1)\right)_{+},\]
where $(\cdot)_{+}$ is an element-wise operator such that $({x})_{+}=\max\{{x},0\}$; see many other efficient projections in \cite{Boyd11}.

%Notice that to apply the optimal scaling lemma, ${\cal B}$ needs not to be a convex set.

{The ADMM updates w.r.t. ${\bf C}$ and ${\bf A}$ are similar; they are relegated to Appendix~\ref{app:ADMM}.
Overall, we solve the subproblems w.r.t. ${\bf A}$, ${\bf B}$, ${\bf C}$ and ${\bf W}$ cyclically as in the last section except that the former three are solved by ADMM.
We should mention that the convergence properties of the described algorithm depend on the type of regularization and the constraints that are added.
The reason is twofold.
First, as previously mentioned, to ensure that every limit point of the solution sequence $\{{\bf A},{\bf B},{\bf C},{\bf W}\}$ is a stationary point of Problem~\eqref{eq:reg_cons_LpFit_0}, each subproblem w.r.t. ${\bf A}$, ${\bf B}$ and ${\bf C}$ needs to be a convex problem admitting a unique solution, which depends on the type of regularization and the constraints.
In addition, the convexity of a subproblem also affects the solution of ADMM - it guarantees that ADMM can attain the optimal solution of that subproblem.}

\subsection{Initialization Approaches}\label{sec:initialization}
IRALS requires initial guesses of ${\bf A}$, ${\bf B}$ and ${\bf C}$.
In practice, several initialization approaches can be considered:

\noindent $\bullet$ First, random initialization is viable.
Since the considered problem is nonconvex, using random initialization may require restarting the algorithm several times from random initial points to attain a good solution,
but it also helps the algorithm to avoid `bad' local minima.

\noindent $\bullet$ Second, the loading factors estimated by algorithms that deal with {the $\ell_2$-norm fitting-based PARAFAC problem in~\eqref{eq:LS-PARAFAC} (or $\ell_2$ PARAFAC for simplicity), e.g., TALS, can be used as starting points.
Algorithms tackling the variants of Problem~\eqref{eq:LS-PARAFAC} with constraints and regularization on the loading factors can also be employed.
This approach is effective when those algorithms are not totally thrown off by the outlying slabs.}

\noindent $\bullet$ Third, {when $I$ is larger than $JK$}, one can first estimate an orthogonal basis ${\bf U}\in\mathbb{R}^{KJ\times R}$ such that ${\cal R}({\bf U})={\cal R}({\bf C}\odot{\bf B})$,
and then apply a Khatri-Rao subspace-based PARAFAC algorithm, such as those in \cite{SOBIUM,lee2013khatri,Veen2006,Veen2001},
on the extracted ${\bf U}$ to get an initial guess of $({\bf B},{\bf C})$.
Khatri-Rao subspace-based initialization is effective with large $I$ since the procedure can `compress' the original tensor substantially\footnote{To be specific, ${\bf U}=({\bf C}\odot{\bf B}){\bm \Theta}$, for some ${\bm \Theta}\in\mathbb{R}^{R\times R}$, can be considered as a compressed tensor which has only $R$ slabs, whereas the original tensor has $I$ slabs. If $R \ll I$, the compressed tensor has many fewer slabs.},
and PARAFAC algorithms empirically work better when the data size is smaller.
When there is no outlying slab, and when both ${\bf C}\odot{\bf B}$ and ${\bf A}$ have full-column rank,
a basis of ${\cal R}({\bf C}\odot{\bf B})$ can be obtained by applying singular value decomposition (SVD) on $\underline{\bf X}^{(3)}$.
Here, since there are outlying columns of $\underline{\bf X}^{(3)}$, we can estimate ${\bf U}$ using robust SVD, which has been intensively studied in the recent literature; see, e.g., \cite{nie2014optimal,xu2010robust}.

\vspace*{-0.1in}
\section{Numerical Results}

In this section,
we first use synthetic data to verify our ideas.
Then, real-data experiments will be presented to show the effectiveness of the proposed algorithmic framework in practice.
The algorithms presented in this section are all implemented in Matlab,
and all simulations and experiments were carried out on a desktop computer with an i7 $3.4$GHz quad-core CPU and 8 GB RAM.

\subsection{Synthetic Data Simulations}
%We first test the proposed algorithms using synthetic data.
In this subsection,
we generate the non-negative loading factors of three-way tensors following the exponential distribution with $\mu=1$.
%In this way, non-negative tensors can be constructed.
%\reminder{From Nikos: Why sparse? In fact they will be dense but with typically small elements.}
The outlier elements are uniformly distributed within zero and one, and then are scaled to satisfy the specified simulation conditions (see below).
To quantify the corruption level, we define the signal-to-outlier {ratio} (SOR) as
\[{\rm SOR(dB)}=10\log_{10}\left(\frac{(1/I)\sum_{i=1}^I\sum_{j=1}^J\sum_{k=1}^K\underline{\bf X}^2(i,j,k)}{(1/|{\cal N}|)\sum_{i\in{\cal N}}\|{\bf O}_i\|_F^2}\right)\]
To benchmark our algorithm, we employ TALS for $\ell_2$ PARAFAC fitting (i.e., Problem~\eqref{eq:LS-PARAFAC}) and the $\ell_1$-norm fitting based PARAFAC ($\ell_1$ PARAFAC) \cite{L1-PARAFAC} with the alternating weighted median filtering realization.
We fix $p=0.5$ throughout this section; our experience is that that the results obtained for different $p \in [0.1,1]$ are qualitatively similar to those obtained for $p=0.5$.
IRALS is stopped when the absolute change of the objective value is less than $10^{-8}$ or the number of iterations reaches $1000$.
%\footnote{{Empirically, using a smaller $p$ may better approximate the $\ell_0$ quasi-norm. Our experience is that setting $p$ to be around $0.5$ balances the performance of the algorithm and the chance of avoiding numerical pitfalls, e.g., `bad' local minima.}},
For IRALS with constraints, we stop the ADMM algorithms for the subproblems when $\|{\bf B}-{\bf B}_1\|_2 + \|{\bf B}-{\bf B}_2\|_2\leq 10^{-3}$ following the guidelines in \cite{Boyd11}.
IRALS and IRALS with constraints are initialized by plain TALS in this subsection.

Table~\ref{tab:SOR} shows the average mean-squared-errors (MSEs) of the estimated ${\bf B}$ and ${\bf C}$ by the algorithms under various SORs;
the runtime performance is also presented in this table.
The MSE of the estimated ${\bf B}$ is defined as
\begin{equation*}
{\rm MSE}=
\min_{ \substack{ \bm{\pi} \in \Pi, \\ c_1,\ldots,c_J \in \{ \pm 1
\}  } }
\frac{1}{K} \sum_{j=1}^J \left\| \frac{ {\bf B}(:,j) }{ \| {\bf B}(:,j) \|_2 } - c_k   \frac{ \hat{\bf B}(:,{\pi_j}) }{ \| \hat{\bf B}(:,{\pi_j}) \|_2 }    \right\|_2^2, %\right\}
\label{eq:MSEdef}
\end{equation*}
where $\Pi$ is the set of all permutations of $\{ 1,2,\ldots,K \}$, and
${\bf B}(:,j)$ and $\hat{{\bf B}}(:,j)$ are the ground truth of the $j$th column of ${\bf B}$ and the corresponding estimate, respectively; the same definition of MSE holds for $\hat{\bf C}$.
%the MSE of the estimated ${\bf C}$ is defined correspondingly.
We see that for $R=5$, IRALS and IRALS with non-negativity constraints (denoted by `IRALS w./ nn') both exhibit much lower MSEs compared to TALS and $\ell_1$ PARAFAC.
When $R=10$, IRALS with non-negativity constraints gives the best MSE performance in general.
In terms of runtime, the unconstrained IRALS and the IRALS with non-negativity constraints are both
faster than $\ell_1$ PARAFAC.
Notably, unconstrained IRALS is more than $50$ times faster than $\ell_1$ PARAFAC in the presented simulations in this table.

%In this case, we do not impose any constraints on the above algorithms.
%Our algorithm is initialized as follows the robust subspace algorithm \cite{nie2014optimal}, as mentioned in the previous section;

Table~\ref{tab:number} shows the MSEs and runtimes versus the number of outlying slabs.
In many cases of this simulation, $\ell_1$ PARAFAC could not yield a reasonable result, and thus it was removed from the comparison.
For the other three algorithms, we see that IRALS and IRALS with non-negativity constraints can yield reasonable estimation of the loading factors even when the number of outlying slabs exceeds a half of the total number of slabs,
but TALS gives very poor estimation in this case.

% Table generated by Excel2LaTeX from sheet 'Sheet2'
\begin{table}[h!]
  \centering
  \caption{The average MSEs of the estimated ${\bf B}$ and ${\bf C}$ by the Algorithms under various SORs; $(I,J,K)=(20,20,20)$; no. of outlying slabs $=6$.}
	
	 \resizebox{8.5cm}{!}{\large

    \begin{tabular}{c|c|c|c|c|c|c}
    \hline
    \hline
    \multicolumn{7}{c}{$R = 5$} \\
    \hline
    \multirow{2}[4]{*}{Algorithm} & \multirow{2}[4]{*}{Measure} & \multicolumn{5}{c}{SOR} \\
\cline{3-7}          &       & -10   & -5    & 0     & 5     & 10 \\
    \hline
    \multirow{2}[4]{*}{TALS} & MSE (dB) & -10.3345 & -13.9955 & -20.6065 & -28.3189 & -34.2078 \\
\cline{2-7}          & TIME (sec.) & 0.0655 & 0.0635 & 0.0579 & 0.0534 & 0.0534 \\
    \hline
    \multirow{2}[4]{*}{L1 PARAFAC} & MSE (dB) & -11.6272 & -19.6811 & -25.499 & -28.156 & -66.0477 \\
\cline{2-7}          & TIME (sec.) & 15.1952 & 12.8005 & 10.9255 & 10.4323 & 10.1826 \\
    \hline
    \multirow{2}[4]{*}{IRALS} & MSE (dB) & \textbf{-28.6011} & -46.3832 & \textbf{-76.4109} & \textbf{-129.469} & -127.115 \\
\cline{2-7}          & TIME (sec.) & 0.2576 & 0.1857 & 0.1496 & 0.1452 & 0.1423 \\
    \hline
    \multirow{2}[4]{*}{IRALS w./ nn} & MSE (dB) & -28.5889 & \textbf{-64.3808} & -73.2943 & -129.468 & \textbf{-127.125} \\
\cline{2-7}          & TIME (sec.) & 8.2756 & 5.7996 & 4.6465 & 4.2905 & 4.1048 \\
    \hline
    \hline
    \multicolumn{7}{c}{$R = 10$} \\
    \hline
    \multirow{2}[4]{*}{Algorithm} & \multirow{2}[4]{*}{Measure} & \multicolumn{5}{c}{SOR} \\
\cline{3-7}          &       & -10   & -5    & 0     & 5     & 10 \\
    \hline
    \multirow{2}[4]{*}{TALS} & MSE (dB) & -9.6438 & -12.9671 & -16.4158 & -24.1415 & -27.1827 \\
\cline{2-7}          & TIME (sec.) & 0.1955 & 0.139 & 0.1174 & 0.0959 & 0.0869 \\
    \hline
    \multirow{2}[4]{*}{L1 PARAFAC} & MSE (dB) & -7.5341 & -10.0898 & -13.4314 & -15.9167 & -17.7212 \\
\cline{2-7}          & TIME (sec.) & 79.3323 & 67.8216 & 52.7992 & 44.2532 & 36.9568 \\
    \hline
    \multirow{2}[4]{*}{IRALS} & MSE (dB) & -19.4927 & -29.2948 & -39.594 & -38.0696 & \textbf{-68.5139} \\
\cline{2-7}          & TIME (sec.) & 0.573 & 0.5305 & 0.4496 & 0.3681 & 0.305 \\
    \hline
    \multirow{2}[4]{*}{IRALS w./ nn} & MSE (dB) & \textbf{-22.5658} & \textbf{-33.7498} & \textbf{-54.5883} & \textbf{-60.9308} & -67.8565 \\
\cline{2-7}          & TIME (sec.) & 16.3145 & 16.4172 & 14.061 & 11.8115 & 9.6166 \\
    \hline
    \hline
    \end{tabular}%
}%
  \label{tab:SOR}%
\end{table}%

% Table generated by Excel2LaTeX from sheet 'Sheet3'
\begin{table}[htbp]
  \centering
  \caption{The average MSEs of the estimated ${\bf B}$ and ${\bf C}$ by the Algorithms versus the number of outlying slabs; $(I,J,K)=(20,20,20)$; SOR $=0$dB; $R=5$.}
    \resizebox{8.5cm}{!}{\large
    \begin{tabular}{c|c|c|c|c|c|c}
    \hline
    \hline
    \multirow{2}[4]{*}{Algorithm} & \multirow{2}[4]{*}{Measure} & \multicolumn{5}{c}{number of outlying slabs} \\
\cline{3-7}          &       & 3     & 5     & 7     & 9     & 11 \\
    \hline
    \multirow{2}[4]{*}{TALS} & MSE (dB) & -15.0081 & -11.4041 & -9.4724 & -8.4138 & -8.0331 \\
\cline{2-7}          & TIME (sec.) & 0.0592 & 0.063 & 0.0635 & 0.0681 & 0.0718 \\
    \hline
    \multirow{2}[4]{*}{IRALS} & MSE (dB) & -30.4836 & -31.3318 & -24.839 & -23.3083 & -23.1527 \\
\cline{2-7}          & TIME (sec.) & 0.1723 & 0.217 & 0.2349 & 0.2474 & 0.2461 \\
    \hline
    \multirow{2}[4]{*}{IRALS w./ nn} & MSE (dB) & \textbf{-37.743} & \textbf{-40.8854 }&\textbf{ -40.8} &\textbf{ -40.4385} & \textbf{-26.3071} \\
\cline{2-7}          & TIME (sec.) & 3.4883 & 4.5614 & 5.3182 & 5.7222 & 5.2961 \\
    \hline
    \hline
    \end{tabular}}%

  \label{tab:number}%
\end{table}%

\begin{figure}
	\centering
		\includegraphics[width=7cm]{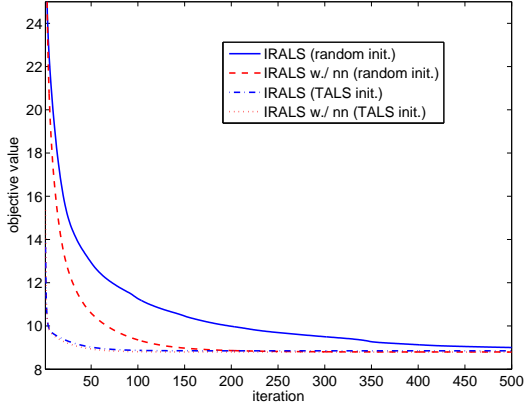}
	\caption{The convergence curves of the objective value when applying IRALS with different initializations and constraints.}
	\label{fig:iteration}
\end{figure}

{Fig.~\ref{fig:iteration} presents the objective values of \eqref{eq:w_PARAFAC} against the iterations when applying IRALS and IRALS with nonnegativity constraints with different initializations.
This simulation is under the settings $I=J=K=20$, $R=5$, and $|{\cal N}|=6$. Each curve is averaged from $100$ trials.
We see that, when using random initialization, the cost function of IRALS with nonnegativity constraints on the loading factors converges much faster than that of IRALS with no constraints.
In addition, using the output of TALS helps the cost functions of both algorithms converge faster.
Specifically, under such an initialization scheme, the objective values given by the algorithms both converge within 100 iterations.}

\subsection{Blind Speech Separation}

In this subsection, we revisit the blind speech separation problem that has been mentioned in Sec.~\ref{sec:motivation}.
We first show a simulation using instantaneously mixed speech sources, where the mixtures follow the signal model in \eqref{eq:instant_mix}.
{The sources are randomly picked from a database that consists of $23$ speech segments;
each source has a length of $3$ second, and is sampled at a rate of $16$KHz.
We use $I=5$ sensors and $R=6$ sources, which poses a challenging under-determined blind separation problem.
Each time frame consists of $200$ samples - this results in $K=239$ time frames (slabs)}.
Spatially and temporally white Gaussian noise is added to the received signals.
Each local covariance of the received signals (i.e., each slab of the PARAFAC model) is calculated using the local sample mean of ${\bf x}(t) ({\bf x}(t))^T$,
and the noise variance is estimated by
\[\hat{\sigma}^2 = \min_{k=1,\ldots,K}~\lambda_{\min}\left(\underline{\bf X}(:,:,k)\right),\]
where $\lambda_{\min}({\bf X})$ denotes the smallest eigenvalue of ${\bf X}$. The estimated noise variance is then removed from the data;
see \cite{lee2013khatri,FuMaHuaSid2015} for details.
The mixing system estimation problem can be formulated as
\[\min_{{\bf A},{\bf C}}~\sum_{k=1}^K\left(\left\|\underline{\bf X}(:,:,k)-{\bf A}{\bf D}_k({\bf C}){\bf A}^T\right\|_F^2+\epsilon\right)^{\frac{p}{2}},\]
and we apply IRALS to the above by treating ${\bf A}{\bf D}_k({\bf C}){\bf A}^T$ as ${\bf A}{\bf D}_k({\bf C}){\bf B}^T$.
In this subsection, we use the Khatri-Rao subspace-based initialization as mentioned in Sec.~\ref{sec:initialization}, since the number of slabs ($K$ in this case) is large.

\begin{figure}
	\centering
		\includegraphics[width=6cm]{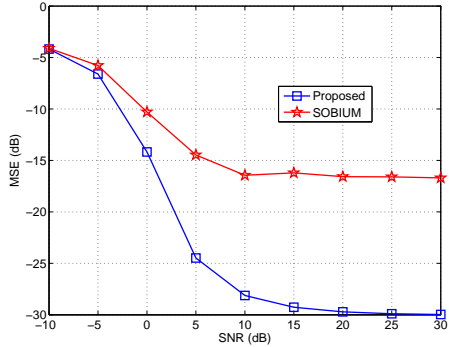}
	\caption{The MSEs of {the estimated mixing systems} obtained by SOBIUM and the proposed algorithm under various SNRs.}
	\label{fig:speech_inst}
\end{figure}

Fig.~\ref{fig:speech_inst} shows the average MSEs of the estimated mixing system obtained by several algorithms; the result is averaged from 100 independent trials.
The benchmarked PARAFAC algorithm is SOBIUM \cite{SOBIUM}, which is known as a state-of-the-art blind source separation algorithm for the under-determined case (i.e., $I<J$).
We see that the proposed algorithm consistently yields around $15$dB lower MSE than that of SOBIUM, which is a significant performance boost.
This phenomenon verifies the existence of (significant) modeling error at some slabs, and also shows the effectiveness of our proposed algorithm.

\begin{figure}
	\centering
		\includegraphics[width=6cm]{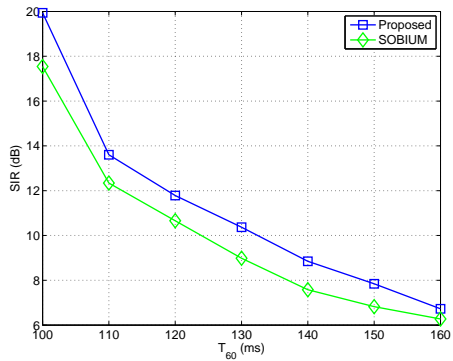}
	\caption{The SIRs obtained by applying SOBIUM and the proposed algorithm to convolutive mixtures under various $T_{60}$'s.}
	\label{fig:speech_conv}
\end{figure}

We also consider the convolutive mixture case, in which the signal model can be represented as
\[{\bf x}(t)=\sum_{\ell=0}^{\ell_{\max}-1}{\bf H}(\ell){\bf s}(t-\ell),\]
where ${\bf x}(t)$ and ${\bf s}(t)$ are defined as before, and ${\bf H}(\ell)$ denotes the mixing system impulse response at time lag $\ell$.
The convolutive mixture model is more realistic, since it captures the multi-path reverberation characteristics of real acoustic environments; but is also far more challenging to deal with, compared to the instantaneous mixture case.
We build up the convolutive mixtures by setting up a simulated room with multiple paths between the speakers and receivers following the image method \cite{Allen1979}.
To separate the sources, we follow the frequency-domain approach \cite{Nion2010,Reilly2005} - the basic idea is to transform the mixtures to the frequency domain,
where the per-frequency (bin) mixtures follow an approximately instantaneous mixing model. Thus, PARAFAC algorithms can be applied at each frequency to obtain the source components at that frequency,
and the time-domain sources can be obtained subsequently using certain post-processing steps, the most critical of which are permutation and scaling alignment across the different frequency bins.
We measure the quality of the unmixed speech signals using the signal-to-interference ratio (SIR) criterion as in \cite{Nion2010,Reilly2005};
higher SIR means better separation performance.
Fig.~\ref{fig:speech_conv} shows the results of using $I=4$ sensors to separate $J=3$ sources; the result is also averaged from $100$ trials with randomly picked sources.
We see that, under different reverberation conditions for the simulated room (a larger $T_{60}$ means a more severe multipath effect, thereby a more challenging environment for speech separation), the proposed algorithm consistently outperforms SOBIUM by around 2dB.

\subsection{Fluorescence Data Analysis}
\begin{figure}
	\centering
		\includegraphics[width=8cm]{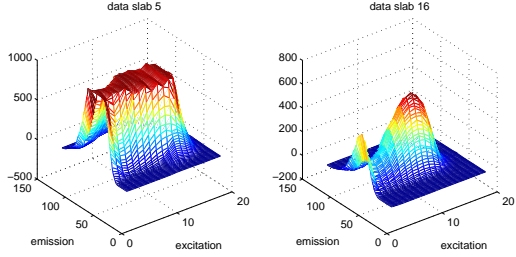}
	\caption{An outlying slab (left) and a relatively clean slab (right) of the Dorrit data.}
	\label{fig:bad_sample}
\end{figure}

In this subsection, we deal with a real fluorescence EEM data set - the Dorrit data that is available online at \url{http://www.models.life.ku.dk/dorrit}.
Our working data set has 116 spectral emissions, 18 excitations, and 27 samples,
which is a tensor with $I=27$, $J= 116$ and $K=18$.
The Dorrit data set is known for containing some badly contaminated slabs, even after pre-processed by some automatic scattering removal algorithm \cite{bahram2006handling},
and there are also some relatively clean samples in this data set; see Fig.~\ref{fig:bad_sample}.
We formulate the problem of estimating the spectral emissions (${\bf B}$) and excitations (${\bf C}$)
as
\begin{align*}
 \min_{{\bf A},{\bf B},{\bf C}}~&\sum_{i=1}^I\left(\left\|\underline{\bf X}(i,:,:)-{\bf B}{\bf D}_i({\bf A}){\bf C}^T\right\|_F^2+\epsilon\right)^{\frac{p}{2}}\\
                                &\quad\quad\quad\quad + \lambda_a\|{\bf A}\|_F^2+\lambda_b\|{\bf T}{\bf B}\|_F^2 + \lambda_c\|{\bf T}{\bf C}\|_F^2\\
                     {\rm s.t.}~&{\bf A}\geq{\bf 0},~{\bf B}\geq{\bf 0},~{\bf C}\geq{\bf 0},
\end{align*}
where ${\bf T}$ is defined in \eqref{eq:smooth_T} with appropriate dimensions.
We add smoothness regularization on ${\bf B}$ and ${\bf C}$ since we know that the emission and the excitation spectra are smooth in practice;
also, non-negativity constraints are added to all three loading factors.
We should point out that adding $\|{\bf A}\|_F^2$ is important;
otherwise, the scaling of ${\bf B}$ and ${\bf C}$ can be `absorbed' by ${\bf A}$, and the smoothness regularization (or, any other scaling-sensitive regularization) may not work.

\begin{figure}
	\centering
		\includegraphics[width=8cm]{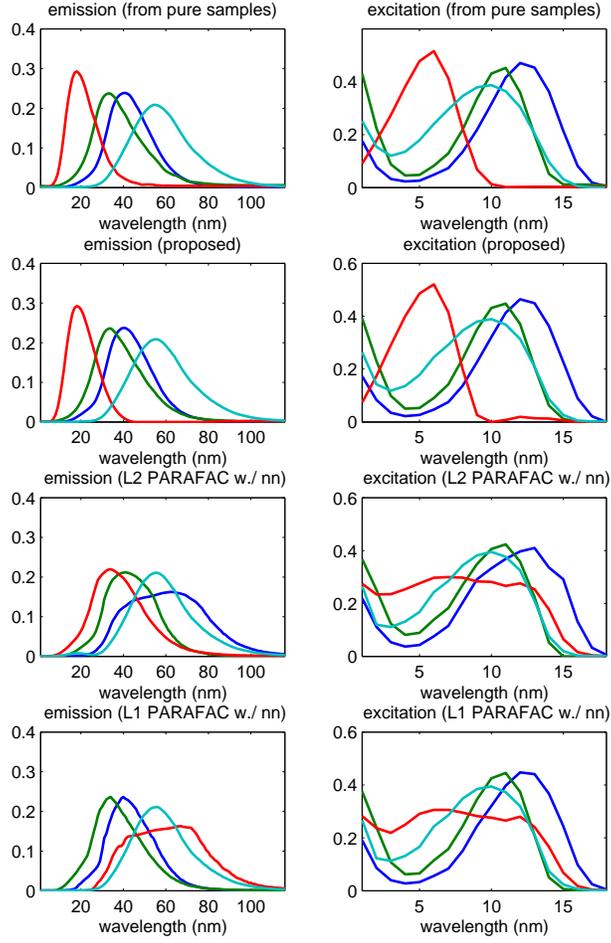}
		\caption{The estimated emission and excitation curves obtained using the proposed algorithm, as well as nonnegativity-constrained $\ell_2$ and $\ell_1$ PARAFAC fitting.}
	\label{fig:emission_dorrit}
\end{figure}

\begin{figure}
	\centering
		\includegraphics[width=8cm,height=3cm]{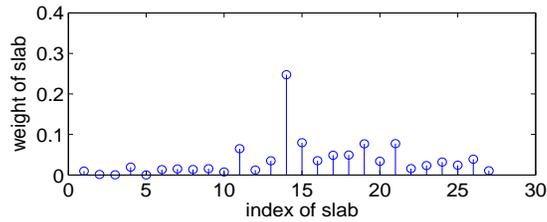}
	\caption{The normalized weights of the samples obtained via IRALS.}
	\label{fig:weights_eem}
\end{figure}

In this experiment, we set $\lambda_b=\lambda_c=10$ and $\lambda_a=10^{-2}$ and $R=4$.
{Here, we use the $\ell_1$ and $\ell_2$ PARAFAC algorithms with nonnegativity constraints as benchmarks, which are both implemented in the $N$-way toolbox \cite{andersson2000n} (available at \url{http://www.models.life.ku.dk/source/nwaytoolbox/}). The result of the nonnegativity-constrained $\ell_2$ PARAFAC algorithm is used to initialize the proposed algorithm.}
The estimated ${\bf B}$ and ${\bf C}$ by the algorithms are shown in Fig.~\ref{fig:emission_dorrit}.
We also provide the emission and excitation spectra obtained from certain `pure samples' containing only a single compound. These pure samples are known from prior studies with this particular dataset, and thus the recovered spectra are believed to be close to the ground truth - see the row tagged as `from pure samples' in Fig.~\ref{fig:emission_dorrit}.
We see that the spectra estimated by the proposed algorithm are visually very similar to those measured from the pure samples.
However, both of the nonnegativity-constrained $\ell_1$ and $\ell_2$ PARAFAC algorithms yield clearly worse results - for both of them, an estimated emission spectrum and an estimated excitation spectrum are highly inconsistent with the results measured from the pure samples.
It is also interesting to observe the weights of the slabs given by the proposed algorithm in Fig.~\ref{fig:weights_eem}.
One can see that the algorithm automatically fully downweights slab $5$, which is consistent with our observation (consistent with domain expert knowledge) that slab 5 is an extreme outlying sample (cf. Fig.~\ref{fig:bad_sample}).
This verifies the effectiveness of our algorithm for joint slab selection and model fitting.

%Particularly, for the spectra estimation, the proposed algorithm resolves all four smooth spectra,
%but the $\ell_1$ model fitting algorithm gives two identical spectra, and a rather erratic spectrum.

%\begin{figure}
	%\centering
		%\includegraphics[width=8cm]{bad_slab.eps}
	%\caption{An outlying slab of the Dorrit data.}
	%\label{fig:bad_sample}
%\end{figure}
%
%\begin{figure}
	%\centering
		%\includegraphics[width=8cm]{good_slab.eps}
	%\caption{A relatively clean slab of the Dorrit data.}
	%\label{fig:clean_sample}
%\end{figure}

%\begin{figure}
	%\centering
		%\includegraphics[width=8cm]{emission.eps}
	%\caption{The estimated spectra by the proposed algorithm and the L1 PARAFAC model fitting.}
	%\label{fig:spectra_dorrit}
%\end{figure}

\subsection{ENRON E-mail Data Mining}
In this subsection, we apply the proposed algorithm on the celebrated ENRON E-mail corpus.
This data set contains the e-mail communications between $184$ persons within 44 months.
Specifically, ${\bf X}(i,j,k)$ denotes the number of e-mails sent by person $i$ to person $j$ within month $k$.
Many studies have been done for mining the social groups out of this data set \cite{bader2006temporal,papalexakis2013k,diesner2005communication}.
In particular, \cite{papalexakis2013k} applied a sparsity-regularized and non-negativity-constrained PARAFAC algorithm on this data set, and some interesting (and interpretable) results have been obtained. In particular, the significant non-zero elements of ${\bf A}(:,r)$ usually correspond to persons with similar `social' positions such as lawyers or executives.

Here, we also aim at mining the social groups out of the ENRON data, while taking data for `outlying months' into consideration.
It is well known that the ENRON company went through a criminal investigation and finally filed for bankruptcy.
Hence, one may conjecture that the e-mail interaction patterns between the social groups might be irregular during the outbreak of the crisis. We fit the data using the following formulation:
\begin{align*}
 \min_{{\bf A},{\bf B},{\bf C}}~&\sum_{k=1}^K\left(\left\|\underline{\bf X}(:,:,k)-{\bf A}{\bf D}_k({\bf C}){\bf B}^T\right\|_F^2+\epsilon\right)^{\frac{p}{2}}\\
                                &\quad\quad\quad\quad \lambda_a f({\bf A})+\lambda_b\|{\bf B}\|_F^2 + \lambda_c\|{\bf C}\|_F^2\\
                     {\rm s.t.}~&{\bf A}\geq{\bf 0},~{\bf B}\geq{\bf 0},~{\bf C}\geq{\bf 0},
\end{align*}
where $f({\bf A})$ is a function that promotes sparsity following the insight in \cite{papalexakis2013k}; $\|{\bf B}\|_F^2$ and $\|{\bf C}\|_F^2$ are added to avoid scaling / counter-scaling issues, as in the previous example.
Notice that here we use an aggressive sparsity promoting function $f({\bf A})$ from \cite{chartrand2014shrinkage}, which itself cannot be put in closed form -- notwithstanding, the proximal operator of $f({\bf A})$ can be written in closed-form, and thus is easy to incorporate into our ADMM framework.
We fit the ENRON data with $R=5$ as in \cite{papalexakis2013k}, and set $\lambda_a= 6.5\times 10^{-2}$, $\lambda_b=\lambda_c=10^{-3}$.
The same pre-processing as in \cite{bader2006temporal,papalexakis2013k} is applied to the non-zero data to compress the dynamic range;
i.e., all the non-zero raw data elements are transformed by an element-wise mapping $x'=\log_2(x)+1$.
{As in the last subsection, the proposed algorithm is initialized by the nonnegativity-constrained $\ell_2$ PARAFAC algorithm.
}

\begin{figure}
	\centering
		\includegraphics[width=8cm,height=3cm]{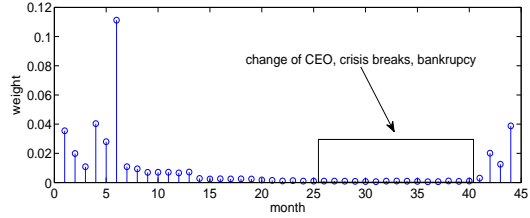}
	\caption{The normalized weights obtained by the proposed algorithm when applied on the ENRON e-mail data.}
	\label{fig:Enron_Weight}
\end{figure}

Table~\ref{tab:ENRON} shows the five social groups mined from the data, corresponding to the non-zero elements in the five columns of ${\bf A}$.
We see that these five groups are quite clean, covering 73 (`important') persons out of 184 in total. %\reminder{From Xiao: I forgot subtracting the number of overlapping people last time, resulting 86 people in the previous version}.
More interestingly,
the algorithm automatically downweights the slabs corresponding to the period when the company was having a crisis - see Fig.~\ref{fig:Enron_Weight}.
This verifies our guess: The interaction pattern during this particular period is not regular, and downweighting these slabs can give us more clean social groups.

% Table generated by Excel2LaTeX from sheet 'Sheet2'
\begin{table*}[htbp]
  \centering
  \caption{Mining the ENRON E-mail corpus using the proposed algorithm.}
	\resizebox{\textwidth}{!}{
       \begin{tabular}{|l|l|l|}
    \hline
    \multicolumn{1}{|c|}{cluster 1 (Legal; 16 persons)} & \multicolumn{1}{c|}{cluster 2 (Excecutive; 18 persons)} & \multicolumn{1}{c|}{cluster 3 (Executive; 25 persons)} \\
    \hline
    Brenda Whitehead, N/A  & David Delainey, CEO ENA and Enron Energy Services & Andy Zipper , VP Enron Online \\
    Dan Hyvl, N/A & Drew Fossum, VP Transwestern Pipeline Company (ETS) & Jeffrey Shankman, President Enron Global Markets \\
    Debra Perlingiere, Legal Specialist ENA Legal & Elizabeth Sager, VP and Asst Legal Counsel ENA Legal & Barry Tycholiz, VP Marketing \\
    Elizabeth Sager, VP and Asst Legal Counsel ENA Legal & James Steffes, VP Government Affairs & Richard Sanders, VP Enron Wholesale Services \\
    Jeff Hodge, Asst General Counsel ENA Legal & Jeff Dasovich, Employee Government Relationship Executive & James Steffes, VP Government Affairs \\
    Kay Mann, Lawyer  & John Lavorato, CEO Enron America & Mark Haedicke, Managing Director ENA Legal \\
    Louise Kitchen, President Enron Online & Kay Mann,  Lawyer  & Greg Whalley, President  \\
    Marie Heard, Senior Legal Specialist ENA Legal & Kevin Presto, VP East Power Trading & Jeff Dasovich, Employee Government Relationship Executive \\
    Mark Haedicke, Managing Director ENA Legal & Margaret Carson,  Employee Corporate and Environmental Policy & Jeffery Skilling, CEO  \\
    Mark Taylor , Manager Financial Trading Group ENA Legal & Mark Haedicke, Managing Director ENA Legal & Vince Kaminski, Manager Risk Management Head \\
    Richard Sanders, VP Enron Wholesale Services & Philip Allen, VP West Desk Gas Trading & Steven Kean, VP Chief of Staff \\
    Sara Shackleton, Employee ENA Legal & Richard Sanders, VP Enron Wholesale Services & Joannie Williamson, Executive Assistant  \\
    Stacy Dickson, Employee ENA Legal & Richard Shapiro , VP Regulatory Affairs & John Arnold, VP Financial Enron Online \\
    Stephanie Panus, Senior Legal Specialist ENA Legal & Sally Beck, COO  & John Lavorato, CEO Enron America \\
    Susan Bailey, Legal Assistant ENA Legal & Shelley Corman, VP Regulatory Affairs & Jonathan McKa, Director Canada Gas Trading \\
    Tana Jones, Employee Financial Trading Group ENA Legal & Steven Kean, VP Chief of Staff & Kenneth Lay, CEO  \\
          & Susan Scott, Employee Transwestern Pipeline Company (ETS) & Liz Taylor, Executive Assistant to Greg Whalley  \\
          & Vince Kaminski, Manager Risk Management Head & Louise Kitchen, President Enron Online \\
\cline{1-2}    \multicolumn{1}{|c|}{cluser 4 (Trading; 12 persons)} & \multicolumn{1}{c|}{cluster 5 (Pipeline; 15 persons)} & Michelle Cash, N/A  \\
\cline{1-2}    Chris Dorland, Manager  & Bill Rapp, N/A  & Mike McConnel, Executive VP Global Markets \\
    Eric Bas, Trader Texas Desk Gas Trading & Darrell Schoolcraft, Employee Gas Control (ETS) & Kevin Presto, VP East Power Trading \\
    Philip Allen, Manager  & Drew Fossum, VP Transwestern Pipeline Company (ETS) & Richard Shapiro, VP Regulatory Affairs \\
    Kam Keiser, Employee Gas & Kevin Hyatt, Director Asset Development TW Pipeline Business (ETS) & Rick Buy, Manager Chief Risk Management Officer \\
    Mark Whitt, Director Marketing & Kimberly Watson, Employee Transwestern Pipeline Company (ETS) & Sally Beck, COO  \\
    Martin Cuilla, Manager Central Desk Gas Trading & Lindy Donoho, Employee Transwestern Pipeline Company (ETS) & Hunter Shively, VP Central Desk Gas Trading \\
    Matthew Lenhart, Analyst West Desk Gas Trading & Lynn Blair, Employee Northern Natural Gas Pipeline (ETS) &  \\
    Michael Grigsby, Director West Desk Gas Trading & Mark McConnell, Employee Transwestern Pipeline Company (ETS) &  \\
    Monique Sanchez, Associate West Desk Gas Trader (EWS) & Michelle Lokay, Admin. Asst. Transwestern Pipeline Company (ETS) &  \\
    Susan Scott, Employee Transwestern Pipeline Company (ETS) & Rod Hayslett, VP Also CFO and Treasurer &  \\
    Jane Tholt, VP West Desk Gas Trading & Shelley Corman, VP Regulatory Affairs &  \\
    Philip Allen, VP West Desk Gas Trading & Stanley Horton, President Enron Gas Pipeline &  \\
          & Susan Scott, Employee Transwestern Pipeline Company (ETS) &  \\
          & Teb Lokey, Manager Regulatory Affairs &  \\
          & Tracy Geaccone, Manager (ETS) &  \\
    \hline
    \end{tabular}%
}%
  \label{tab:ENRON}%
\end{table*}%

%\reminder{From Nikos: What is the effect of changing $p$ to $0.1$ or $0.9$ for example? Does this have a significant effect on the results obtained? We should mention something about this.}

\vspace*{-0.1in}
\section{Conclusion}
In this work, we considered the problem of low-rank tensor decomposition in the presence of outlying slabs.
Several practical motivating applications have been introduced.
A conjugate augmented optimization framework has been proposed to deal with the formulated $\ell_p$ minimization-based factorization problem.
The proposed algorithm features similar complexity as the classic TALS algorithm that is not robust to outlying slabs.
Regularized and constrained optimization has also been considered by employing an ADMM update scheme.
Simulations using synthetic data and experiments using real data have shown that the proposed approach is promising in different pertinent applications such as blind speech separation, fluorescence data spectroscopy, and social network mining.

%---------------------------------------------------------------------------
%\clearpage
\appendix

\ifplainver
    \section*{Appendix}
    \renewcommand{\thesubsection}{\Alph{subsection}}
\else
    \section{Appendix}
\fi

\subsection{Proof of Claim~\ref{thm:L0PARAFAC}}\label{app:identifiability}
Consider a feasible solution $(\tilde{\bf A},\tilde{\bf B},\tilde{\bf C})$, where $\tilde{\bf A}({\cal N}_c,:)={\bf A}({\cal N}_c,:){\bm \Pi}{\bm \Delta}_a$, $\tilde{\bf B}={\bf B}{\bm \Pi}{\bm \Delta}_b$, and $\tilde{\bf C}={\bf C}{\bm \Pi}{\bm \Delta}_c$.
%Our goal is to show that there does not exists a solution which admits a smaller objective value than that of $({\bf A}^\star,{\bf B}^\star,{\bf C}^\star)$.
Consequently, it can be seen that for all $i\in{\cal N}_c$ we have
\[{\cal I}\left(\left\| \underline{{\bf  X}}^{(3)}(:,i) - (\tilde{\bf  C}\odot\tilde{\bf  B})\tilde{\bf  A}(i,:)^T \right\|_2 \right)= 0.\]
Hence, the optimal value of the cost function satisfies
\[v_{\rm min} \leq I - |{\cal N}_c| \leq \frac{I-c}{2}.\]

Now, we show that there is no other solution that leads to a smaller objective value.
Suppose that there exists an index set ${\cal S}\subseteq{\cal N}_c$ such that (some of)
the slabs indexed by $i\in{\cal S}\bigcup {\cal N}$ constitute a PARAFAC model whose loading matrices do not contain ${\bf B}{\bm \Pi}{\bm \Delta}_b$ or ${\bf C}{\bm \Pi}{\bm \Delta}_c$.
We show that $|{\cal S}|<c$.
In fact, if $|{\cal S}|\geq c$, then, with probability one, the slabs that belong to ${\cal S}$ can only be decomposed using ${\bf B}$, ${\bf C}$ and ${\bf A}({\cal S},:)$ with a common column permutation and scaling.
The reason is as follows. By the assumption that ${\bf A}$ is drawn from some absolutely continuous distribution,
we see that $k_{{\bf A}({\cal S},:)}=\min\{|{\cal S}|,R\}\geq \min\{c,R\}= c$ holds with probability one, and thus $k_{{\bf A}({\cal S},:)}+\min\{J,R\} + \min\{K,R\}\geq 2R+2$ holds almost surely.
%The reason is that $c \leq \min\{I,R\}$ implies that \eqref{eq:Kruskal_random} holds.
Hence, by the uniqueness condition mentioned in \eqref{eq:Kruskal_random}, the PARAFAC decomposition of $\underline{\bf X}({\cal S},:,:)$ is essentially unique with probability one.
Thus, it can be seen that if the solution to Problem~\eqref{eq:L0Fit} does not satisfy ${\bf B}^\star={\bf B}{\bm \Pi}{\bm \Delta}_b$,  and ${\bf C}^\star={\bf C}{\bm \Pi}{\bm \Delta}_c$,
we must have
\[{v}_{\rm min} \geq I - \left|{\cal N}\bigcup{\cal S}\right| > I - \frac{I+c}{2}=\frac{I-c}{2},\]
where we have used the fact that $|{\cal N}\bigcup{\cal S}|<(I+c)/2$.

It remains to show that ${\bf A}^\star({\cal N}_c,:)={\bf A}({\cal N}_c,:){\bm \Pi}{\bm \Delta}_a$.
In fact, given that the optimal solution satisfies ${\bf B}^\star={\bf B}{\bm \Pi}{\bm \Delta}_b$, and ${\bf C}^\star={\bf C}{\bm \Pi}{\bm \Delta}_c$,
the optimal ${\bf A}^\star$ should be able to make
\begin{equation}\label{eq:A_Nc}
{\cal I}(\| \underline{{\bf  X}}^{(3)}(:,i) - ({\bf  C}^\star\odot {\bf  B}^\star){\bf  A}^\star(i,:)^T \|_2)= 0,
\end{equation}
for as many as possible $i$'s.
%is ${\bf A}^\star({\cal N}_c,:)={\bf A}({\cal N}_c,:){\bm \Pi}{\bm \Delta}_a$.
For $i\in{\cal N}_c$, we conclude ${\bf A}^\star({\cal N}_c,:)={\bf A}({\cal N}_c,:){\bm \Pi}{\bm \Delta}_a$.
The reason, again, lies in the uniqueness result in \eqref{eq:Kruskal_random}: Since $|{\cal N}_c|\geq (I+c)/2 \geq c$,
we have ${k}_{{\bf A}({\cal N}_c,:)}\geq c$ with probability one, since ${\bf A}$ is drawn from an absolutely continuous distribution over $\mathbb{R}^{I\times R}$. Hence, ${k}_{{\bf A}({\cal N}_c,:)}+k_{\bf B}+k_{\bf C}\geq 2R+2$ holds with probability one. Consequently, the PARAFAC decomposition of $\underline{\bf X}({\cal N}_c,:,:)$ is essentially unique. This implies ${\bf A}^\star({\cal N}_c,:)={\bf A}({\cal N}_c,:){\bm \Pi}{\bm \Delta}_a$.
%is the unique solution to achieve \eqref{eq:A_Nc} for all $i\in{\cal N}_c$.

\vspace*{-0.1in}
\subsection{Proof of Claim~\ref{prop:convergence}}\label{app:claim2}

To relate the stationary points of Problem~\eqref{eq:w_PARAFAC} to the stationary points of Problem~\eqref{eq:LpFit},
let us denote the cost functions of Problem~\eqref{eq:LpFit} and Problem~\eqref{eq:w_PARAFAC} as
$\Psi_1({\bf A},{\bf B},{\bf C})$
and $\Psi_2({\bf A},{\bf B},{\bf C},{\bf W})$, respectively.
We see that
\[ \Psi_1({\bf A},{\bf B},{\bf C}) = \min_{w_1,\ldots,w_I\geq{0}}~\Psi_2({\bf A},{\bf B},{\bf C},\{w_i\}_{i=1}^I).\]
Let us consider $({\bf A}^\star,{\bf B}^\star,{\bf C}^\star,\{w_i^\star\}_{i=1}^I)$ as a stationary point of Problem~\eqref{eq:w_PARAFAC}.
Following Lemma~\ref{lem:conjugate}, a direct observation is that
\begin{equation}\label{eq:one_quals_to_two}
\Psi_1({\bf A}^\star,{\bf B}^\star,{\bf C}^\star) = \Psi_2({\bf A}^\star,{\bf B}^\star,{\bf C}^\star,\{w_i^\star\}_{i=1}^I),
\end{equation}
since $\Psi_2({\bf A},{\bf B},{\bf C},\{w_i\}_{i=1}^I)$
has a unique stationary point w.r.t. $\{w_i\}_{i=1}^I$ on the interior of the nonnegative orthant, which is the optimal solution w.r.t. $\{w_i\}_{i=1}^I$.
Hence, one can see that ${\bf A}^\star,{\bf B}^\star,{\bf C}^\star$
is also a stationary point of $ \Psi_1({\bf A},{\bf B},{\bf C}) $.
In fact, taking ${\bf A}^\star$ as an example, we see that, following \eqref{eq:one_quals_to_two},
\begin{align*}
&{\rm Tr}\left(\nabla_{\bf A}\Psi_2({\bf A}^\star,{\bf B}^\star,{\bf C}^\star,\{w_i^\star\}_{i=1}^I)^T({\bf A}-{\bf A}^\star)\right)\leq {\bm 0}\\
&\Rightarrow {\rm Tr}\left(\nabla_{\bf A} \Psi_1({\bf A}^\star,{\bf B}^\star,{\bf C}^\star)^T({\bf A}-{\bf A}^\star)\right)\leq {\bm 0},
\end{align*}
which implies that ${\bf A}$ is also a stationary point of Problem~\eqref{eq:LpFit}.
The same proof applies to ${\bf B}$ and ${\bf C}$.

\vspace*{-0.2in}
\subsection{ADMM Updates w.r.t. {\bf C} and ${\bf A}$}\label{app:ADMM}
Now, let us consider the update of ${\bf C}$:
\begin{equation}
\begin{aligned}
      \min_{\bf C}&\quad\frac{1}{2}\left\|({\bf  I}\otimes{\bf  W})\underline{\bf  X}^{(2)} -\left({\bf  B}\odot{\bf  W}{\bf  A}\right){\bf C}^T\right\|_F^2 + \lambda_c h({\bf C})\\
			{\rm s.t.}&\quad{\bf C}\geq{\bf 0}.
\end{aligned}
\end{equation}
By applying the same structure of ADMM, we come up with
\begin{subequations}
\begin{align}
 {\bf C}_1^T &:= \left({\bf B}^T{\bf B}\circledast{\bf A}^T{\bf W}^2{\bf A} + \rho{\bf I}\right)^{-1} \times \nonumber\\
 &\left(({\bf B}\odot{\bf W}^2{\bf A})^T\underline{\bf  X}^{(2)}+\rho({\bf C}+{\bf V}_1)^T\right)\\
 {\bf C}_2& :=\arg\min_{{\bf C}_2} \lambda_c h({\bf C}_2) + \frac{\rho}{2}\|{\bf C}_1 - {\bf C} + {\bf V}_2\|_F^2\\
 {\bf C}& :=\left(\frac{1}{2}({\bf C}_2 - {\bf V}_2 + {\bf C}_1 - {\bf V}_1)\right)_{\cal C}\\
 {\bf V}_1& := {\bf V}_1 + {\bf C} - {\bf C}_1\\
 {\bf V}_2& := {\bf V}_2 + {\bf C} - {\bf C}_2.
 \end{align}
\end{subequations}

The update w.r.t. ${\bf A}$ is even simpler:
\begin{subequations}
\begin{align}
 {\bf A}_1^T &:= \left({\bf C}^T{\bf C}\circledast{\bf B}^T{\bf B} + \rho{\bf I}\right)^{-1}\left(({\bf C}\odot{\bf B})^T\underline{\bf  X}^{(3)}+\rho({\bf A}+{\bf Z}_1)^T\right)\\
 {\bf A}_2& :=\arg\min_{{\bf A}_2} \lambda_a f({\bf A}_2) + \frac{\rho}{2}\|{\bf A} - {\bf A}_1 + {\bf Z}_2\|_F^2\\
 {\bf A}& =\left(\frac{1}{2}({\bf A}_1 - {\bf Z}_1 + {\bf A}_2 - {\bf Z}_2)\right)_{\cal A}\\
  {\bf Z}_1& := {\bf Z}_1 + {\bf A} - {\bf A}_1\\
	 {\bf Z}_2& := {\bf Z}_2 + {\bf A} - {\bf A}_2.
 \end{align}
\end{subequations}

\vspace*{-0.2in}
\bibliography{refs_thesis}

\end{document}